**An uncertainty principle for neural coding: Conjugate representations of position and velocity are mapped onto firing rates and co-firing rates of neural spike trains**

Ryan Grgurich and Hugh T. Blair

Psychology Department, UCLA, Los Angeles, CA

**Abstract.** The hippocampal system contains neural populations that encode an animal's position and velocity as it navigates through space. Here, we show that such populations can embed two codes within their spike trains: a *firing rate code* ($R$) conveyed by within-cell spike intervals, and a *co-firing rate code* ($\dot{R}$) conveyed by between-cell spike intervals. These two codes behave as *conjugates* of one another, obeying an analog of the uncertainty principle from physics: information conveyed in $R$ comes at the expense of information in $\dot{R}$, and vice versa. An exception to this trade-off occurs when spike trains encode a pair of conjugate variables, such as position and velocity, which do not compete for capacity across $R$ and $\dot{R}$. To illustrate this, we describe two biologically inspired methods for decoding $R$ and $\dot{R}$, referred to as *sigma* and *sigma-chi* decoding, respectively. Simulations of head direction (HD) and grid cells show that if firing rates are tuned for position (but not velocity), then position is recovered by sigma decoding, whereas velocity is recovered by sigma-chi decoding. Conversely, simulations of oscillatory interference among theta-modulated "speed cells" show that if co-firing rates are tuned for position (but not velocity), then position is recovered by sigma-chi decoding, whereas velocity is recovered by sigma decoding. Between these two extremes, information about both variables can be distributed across both channels, and partially recovered by both decoders. These results suggest that neurons with different spatial and temporal tuning properties—such as speed versus grid cells—might not encode different information, but rather, distribute similar information about position and velocity in different ways across $R$ and $\dot{R}$. Such conjugate coding of position and velocity may influence how hippocampal populations are interconnected to form functional circuits, and how biological neurons integrate their inputs to decode information from firing rates and spike correlations.

## INTRODUCTION

The rodent hippocampal system contains populations of neurons that encode an animal's position and velocity as it navigates through space. In the literature, these populations are named for variables that modulate their firing rates: "head-direction" (HD) cell firing rates are tuned for the angular position of the head (Taube et al., 1990), "grid" cell firing rates are periodically tuned for the animal's spatial position (Hafting et al., 2005), "place" cell firing rates are non-periodically tuned for the animal's spatial position (O'Keefe & Dostrovsky, 1971), "border" (or "boundary") cell firing rates increase (or decrease) near environmental boundaries (Solstad et al., 2008; Savelli et al., 2008; Lever et al., 2009), "speed" cell firing rates increase (or decrease) in proportion with the animal's running speed (Kropf et al., 2015; Hinman et al., 2016; Gois & Tort, 2018), and "theta" cell firing rates are temporally modulated by 4-12 Hz theta oscillations. Although these monikers accurately describe how the firing rates of individual neurons are tuned, they may not fully describe the information that such neurons encode at the population level, because populations of spiking neurons can encode information not only in their firing rates, but in other ways as well.

Position-tuned neurons (such as place and grid cells) exhibit oscillatory modulation of their spike trains by theta rhythm. Information about the animal's position can be decoded not only from the firing rates of these neurons (Wilson and McNaughton, 1993), but also from the phases at which spikes occur relative to theta rhythm in the local field potential (O'Keefe and Recce, 1993; Skaggs et al., 1996; Jensen and Lisman, 2000; Hafting et al., 2008; Climer et al., 2013; Jeewajee et al., 2013). Hence, spatially tuned neurons can encode the animal's position in two different ways: in their firing rates and firing phases. These two codes map the animal's position into different representational spaces. The firing rate code maps position into a space of "population vectors" where each dimension measures time intervals between pairs of spikes that are both fired by the *same* neuron (because a neuron's firing rate is simply the inverse of its mean interspike interval). By contrast, the phase code maps position into a representational space where each dimension measures time intervals (normalized by the theta cycle period) between pairs of spikes that are fired by two *different* neurons (for example, between a place or grid cell and a neuron that spikes in synchrony with the theta LFP). If neural populations can simultaneously encode information in different ways, then how should information about the world be distributed among different coding channels?

Here, we shall propose how neural populations divide information between two coding channels, which shall be referred to as *firing rates* versus *co-firing rates*. It is worthwhile to further explain what is meant here by the term "co-firing rate." A traditional firing rate code may be considered an example of a *within-cell spike code* that maps information onto intervals between pairs of spikes fired by the same neuron. By contrast, a phase code may be considered an example of a *between-cell spike code* that maps information onto intervals between pairs of spikes fired by different neurons. This distinction between within- versus between-cell spike codes is orthogonal to the distinction between "rate" and "time" codes. In addition to being a within-cell spike code, a traditional firing rate code is also a "rate" code (hence the name, firing "rate"), because firing rates are derived by averaging spike intervals over time, and this discards information about precise temporal sequences of spikes. Alternatively, consider a code that represents information using precise temporal sequences of spikes fired by a single neuron; this would be an example of a within-cell spike code that is also a "time" code. Hence, standard firing rate codes are a particular subset of within-cell spike codes, namely, those which map information onto time-averaged within-cell spike intervals (rather than onto temporal patterns of within-cell spike intervals). We may analogously define "co-firing rate codes" to be the subset of between-cell spike codes that map information onto time-averaged between-cell spike intervals, rather than onto temporal patterns of between-cell spike intervals. Note that standard firing rates are an unsigned quantity (there is no such thing as a negative firing rate), but co-firing rates are a signed quantity, because a distinction can be drawn between the mean interval at which a spike from neuron A follows a spike from neuron B (positive co-firing rate) and the mean interval at which a spike from neuron B follows a spike from neuron A (negative co-firing rate). In summary, we define a "rate code" broadly to be any code that maps information onto *time-averaged* spike intervals, rather than onto temporal patterns of spike intervals. The term "firing rate" shall henceforth refer to time-averaged measurements of within-cell spike intervals, and the term "co-firing rate" shall refer to time-averaged measurements of between-cell spike intervals.

We propose below that firing rates and co-firing rates behave as *conjugate coding channels*, and consequently, their information content is regulated by principles that mirror the well-known "uncertainty principle" from physics. The uncertainty principle states that the more accurately we know a particle's position, the less accurately we can know its momentum (which is identical to velocity, after normalizing for mass), and vice versa. This trade-off arises because positon and velocity are *conjugate variables*, related to one another via the Fourier transform. Simulations presented below show that a neural population can simultaneously embed two codes within its spike trains: a *firing rate code* ($R$) derived from within-cell spike intervals, and a *co-firing rate code* ($\dot{R}$) derived from between-cell spike intervals. It is postulated that $R$ and $\dot{R}$ behave as conjugates of one another, and thus obey an *uncertainty principle for neural coding*: the more information is encoded by $R$, the less information can be encoded by $\dot{R}$ (and vice versa). Hence, the firing rate and co-firing rate channels together cannot convey more information than either channel alone. But a special case arises when neurons simultaneously encode a variable, $q$, and its conjugate, $\dot{q} = dq/dt$ (for example, position and velocity). In such cases, conjugacy between $R$ and $\dot{R}$ complements conjugacy between $q$ and $\dot{q}$, and states of the phase space $q \times \dot{q}$ can be encoded simultaneously in firing rates and co-firing rates, without competition for capacity across channels. The conjugate relationship between $R$ and $\dot{R}$ thus confers limitations as well as advantages for packaging information about the world into neural spike trains. We shall express these constraints as a set of formal mathematical postulates (yet to be proven), and discuss how these constraints might influence neural representations of position and velocity in the hippocampal system. We also discuss how conjugate coding might impact the computations that biological neurons perform upon their spike train inputs, enabling them to extract information from both firing rates and co-firing rates.

**RESULTS**

        All simulations presented here shall use simulated spike trains, artificially generated from real behavioral data (no spike trains recorded from biological neurons are used). A single neuron's spike train shall be represented by a neural response function stored as a series of binary values sampled synchronously at discrete time points,

$$\mathbf{s}_n = \{s_n[t_0], s_n[t_1], \cdots, s_n[t_k], \cdots, s_n[t_K]\}, \qquad (\text{Eq.1})$$

where $s_n[t_k] = 1$ if neuron $n$ fires a spike at time $t_k$, and $s_n[t_k] = 0$ otherwise. A fixed sampling rate of 1 KHz shall be used for all simulations. Spike trains generated by a population of $N$ neurons are stored in a spike response matrix:

$$\mathbf{S} = \begin{bmatrix} \mathbf{s}_1 \\ \mathbf{s}_2 \\ \vdots \\ \mathbf{s}_N \end{bmatrix} = \{S_0, S_1, \ldots, S_k, \ldots, S_K\}, \qquad (\text{Eq. 2})$$

where each row, $\mathbf{s}_n$, contains the neural response function of a different neuron, and each column, $S_k = \{s_1[t_k], s_2[t_k], \ldots, s_N[t_k]\}^T$, contains a population vector of binary values flagging which neurons fired a spike at time $t_k$.

**Computing angular velocity by differentiating HD cell spike trains**

        In this section, spike trains of HD cells are simulated to encode an animal's angular head position—but not angular head velocity—in their firing rates. It is shown that head angle can be recovered from the HD cell population's firing rates by process referred to as *sigma decoding*. It

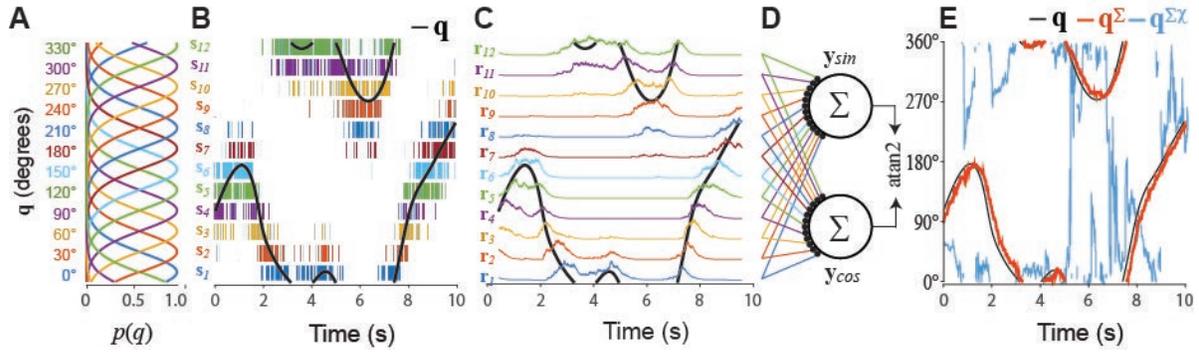

**Figure 1.** *Sigma decoder recovers head angle from HD cell firing rates*. A) Von Mises tuning functions were used to simulate spike trains of 12 HD cells. B) Black line shows 10 s sample of head angle data, *q*; rasters show spike trains of 12 HD cells. C) Head angle data (black line) superimposed over colored lines showing HD cell firing rates derived by Eq. 5 ($\tau_u$=200 ms). D) Sigma neurons sum simulated HD cell firing rates to decode sine and cosine components of the head angle. E) Arctangent function recovers the estimated head angle, $\mathbf{q}^{\Sigma}$, from its sine and cosine components; head angle cannot be recovered by the sigma-chi decoder, $\mathbf{q}^{\Sigma\chi}$.

is further shown that even if individual neurons do not encode information about angular velocity in their firing rates, velocity information can nonetheless be recovered from the HD cell population's co-firing rates by a process referred to as *sigma-chi decoding*.

*Simulation of HD cell spike trains*

Simulated HD cell spike trains were generated from head angle data obtained while a behaving rat foraged freely for food pellets in a cylindrical arena (see Methods). Tracking data was upsampled to 1 KHz to match the spike train simulation rate, yielding a time series of head azimuth angles, $\mathbf{q} = \{q_0, q_1, \cdots, q_k, \cdots, q_K\}$. HD cell tuning curves were simulated by a Von Mises distribution, so that the probability of HD cell $n$ firing a spike at time $t_k$ was given by:

$$p_n[t_k] = R_{max}\frac{e^{\kappa\cos(q_k-\hat{q}_n)}}{2\pi I_0(\kappa)}\,dt,\qquad\text{(Eq. 3)}$$

where $\hat{q}_n$ is the preferred firing direction for neuron $n$, $R_{max}$ is the peak firing rate of the HD cell in Hz, $\kappa$ is a concentration parameter which regulates the HD tuning width, $I_0(\kappa)$ is the modified Bessel function of order 0, and $dt$=1 ms is the time bin resolution. Note that parameters $R_{max}$ and $\kappa$ are not indexed by $n$, because all HD cells shared the same values of these parameters. Fig. 1A shows HD tuning curves for $N = 12$ simulated HD cells with $R_{max} = 100$ Hz and $\kappa = 0.5$. At each time step, the value of $p_n[t_k]$ was used as a binary threshold on the output of pseudorandom number generator, to stochastically determine whether HD cell $n$ fired a spike at time $t_k$. The binary values generated by this stochastic process filled a spike response matrix (Eq. 2) for the HD cell population. Fig. 1B shows spike rasters for 12 simulated HD cells during a 10 s segment of head angle data.

*Sigma decoding: Recovering head angle from HD cell firing rates*

The head angle can be recovered from simulated HD cell spike trains using a standard vector summation method for decoding neural firing rates. This method—henceforth referred to as *sigma decoding*—involves three steps, which can be written as a sequence of mapping functions: $\mathbf{S} \rightarrow \mathbf{R} \rightarrow \mathbf{Y} \rightarrow \mathbf{q}^{\Sigma}$. The first step, $\mathbf{S} \rightarrow \mathbf{R}$, converts spike trains into firing rates via a process that mimics temporal integration of spike trains at synapses. The second step, $\mathbf{R} \rightarrow \mathbf{Y}$, computes weighted sums of firing rates to obtain estimates for the sine and cosine components of the head angle. The third step, $\mathbf{Y} \rightarrow \mathbf{q}^{\Sigma}$, applies the arctangent function to estimate the head angle from its sine and cosine components. We shall write $\mathbf{q}^{\Sigma}$ to denote the time series of estimated head angles generated by the sigma decoder.

Step 1: Converting spike trains into firing rates. HD cell spike trains were converted into firing rates via a process intended to mimic temporal integration at synapses. The input to this process is a spike response matrix $\mathbf{S}$ (Eq. 2), and the output is a *firing rate response matrix*:

$$\mathbf{R} = \begin{bmatrix} \mathbf{r}_1 \\ \mathbf{r}_2 \\ \vdots \\ \mathbf{r}_N \end{bmatrix} = \{R_0, R_1, \dots, R_k, \dots, R_K\}. \tag{Eq. 4}$$

Each column of this matrix, $R_k = \{r_1[t_k], r_2[t_k], \dots, r_N[t_k]\}^T$, is a population vector containing instantaneous HD cell firing rates measured at time $t_k$. Each row, $\mathbf{r}_n = \{r_n[t_0], r_n[t_1], \cdots, r_n[t_K]\}$, is a time series of firing rate estimates for a single HD cell obtained by convolving its spike train with an exponential decay kernel,

$$\mathbf{r}_n = \mathbf{s}_n * \mathbf{u}. \tag{Eq. 5}$$

Individual elements of the decay kernel, $\mathbf{u} = \{u_0, u_1, \dots, u_l, \dots, u_{\infty}\}$, are given by

$$u_l = e^{-l/\tau_u}, \tag{Eq. 6}$$

where $\tau_u$ is the decay constant, and $l$ indexes time bins with positive offsets from $p_n[t_k]$ (kernel weights were uniformly zero for negative time offsets). Fig. 1C illustrates examples of simulated HD cell firing rates with $\tau_u = 200$ ms.

Step 2: Converting firing rates into angle components. Firing rates were converted into sine and cosine components of the head angle via spatial integration of synaptic inputs to a pair of sigma units (Fig 1D) that decoded $\sin(\mathbf{q})$ and $\cos(\mathbf{q})$. Activation of these two sigma units can be represented by a response matrix, $\mathbf{Y} = \{\mathbf{y}_{sin}, \mathbf{y}_{cos}\}^T = \{Y_0, Y_1, \dots, Y_k, \dots, Y_K\}$, where the two rows $\mathbf{y}_{sin}$ and $\mathbf{y}_{cos}$ are time series of decoded estimates for $\sin(\mathbf{q})$ and $\cos(\mathbf{q})$, and each column $Y_k = \{sin_k, cos_k\}^T$ contains the estimated sine and cosine of the decoded head angle at time $t_k$. The output from decoder neuron $m$ was computed as a weighted sum of its firing rate inputs:

$$y_m[t_k] = \sum_{n=1}^{N} w_{m,n} r_m[t_k]$$

$$\tag{Eq. 7}$$

where $w_{m,n}$ is a synaptic weight assigned to the $n^{th}$ spike train input received by decoder neuron $m$. In most simulations, input weights were $w_{m,n} = \sin(2\pi n/N)$ or $w_{m,n} = \cos(2\pi n/N)$ for the sine versus cosine decoder neurons, respectively. But in some simulations (see below, "Conjugacy of firing rates and co-firing rates"), the vector $w_{m,n}$ was derived via the pseudoinverse to obtain an optimal fit between the decoded versus actual head angle data.

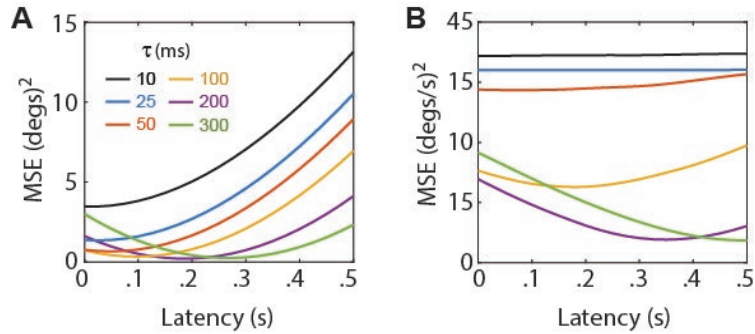

**Figure 2.** *Accuracy-latency tradeoff for spike train decoding.* A) Mean-squared error (MSE; y-axis) for decoding angular head position from simulated HD cell firing rates at varying latencies (x-axis); larger decay constants (τ) improve decoding accuracy at a cost of increased latency. B) MSE for decoding angular velocity from simulated HD cell co-firing rates at varying latencies; optimal latencies are approximately doubled from those in 'A' because the sigma-chi decoder of co-firing rates uses two temporal integration steps.

<u>Recovering head angle via the arctangent function</u>. The final decoding step, $\mathbf{Y} \rightarrow \mathbf{q}^{\Sigma}$, was performed by taking the arctangent of decoder neuron outputs at each time step:

$$q_k^{\Sigma} = \text{atan2}(sin_k, cos_k). \qquad \text{(Eq. 8)}$$

Figure 1E shows that $\mathbf{q}^{\Sigma}$ (red line) provides an accurate estimate of $\mathbf{q}$ (black line), except that $\mathbf{q}^{\Sigma}$ is delayed in time with respect to $\mathbf{q}$ by a latency that is approximately equal to the exponential decay constant ($\tau_u = 200$ ms for the simulation shown). The dynamics of the animal's head turning set an upper bound on $\tau_u$, because the decay constant must not be set so large that $\mathbf{q}$ can undergo large changes within the span of the integration time window. Below this upper bound, there is a tradeoff between accuracy and latency of the decoded position signal, such that increasing $\tau_u$ to integrate over longer time periods improves the accuracy of $\mathbf{q}^{\Sigma}$ (since a larger time window contains more spikes and thus more information), at a cost of delaying the decoded signal in time (Fig. 2A). This type of latency shift can be eliminated if inputs to the decoder are modulated by velocity as well as position (Eliasmith, 2005), but this trick will not be utilized here, because our purpose is to demonstrate how velocity information can be recovered from co-firing rates in the case where individual firing rates exclusively encode position, and contain no information about velocity.

*Sigma-chi decoding: Recovering angular velocity from HD cell co-firing rates*

Simulated HD cell firing rates were modulated only by the angular head position, $\mathbf{q}$, and not by the angular head velocity $\dot{\mathbf{q}}$ (Eq. 3). Nonetheless, $\dot{\mathbf{q}}$ can be decoded from HD cell spike trains using a process we shall refer to as *sigma-chi decoding*, which involves a sequence of three steps: $\mathbf{S} \rightarrow \vec{\mathbf{X}} \rightarrow \dot{\mathbf{R}} \rightarrow \dot{\mathbf{q}}^{\Sigma\chi}$. The first step, $\mathbf{S} \rightarrow \vec{\mathbf{X}}$, performs nonlinear integration on pairs of spike trains to derive a quantity we call the *chi rate*. The second step, $\vec{\mathbf{X}} \rightarrow \dot{\mathbf{R}}$, pools chi rates to extract a vector of *co-firing rates* that form a frequency domain representation of angular velocity, encoded by a population of neurons called *sigma-chi units*. The third step, $\dot{\mathbf{R}} \rightarrow \dot{\mathbf{q}}^{\Sigma\chi}$, decodes angular head velocity from sigma-chi units via vector summation of co-firing rates. We write $\dot{\mathbf{q}}^{\Sigma\chi}$ to denote the time series of angular head velocities recovered by the sigma decoder.

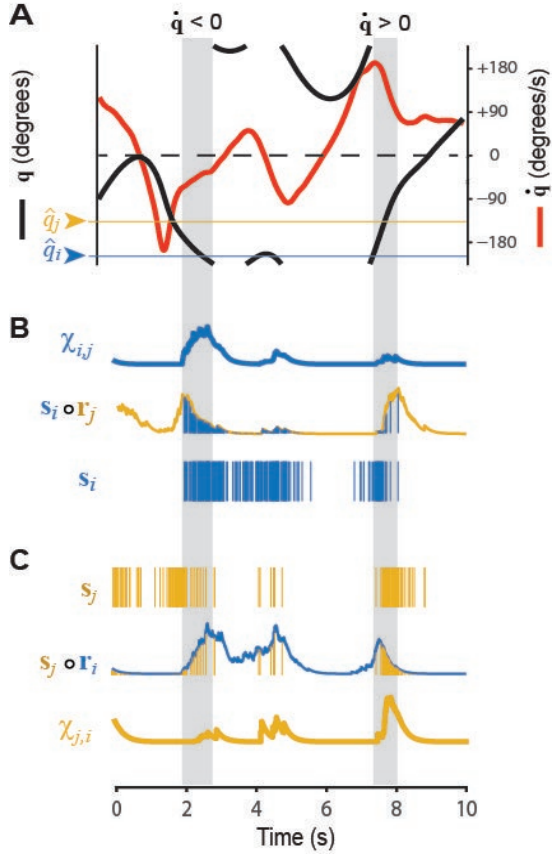

**Figure 3.** *Chi rates for a pair of HD cells.* A) Black line shows 10 s sample of the head angle *q*, red line shows angular head velocity $\dot{q}$ during the same period; colored lines show preferred firing directions for simulated HD cells *i* and *j* (which are the same as HD cells 1 and 3, respectively, from Fig. 1). B,C) Dependency of chi rates $\chi_{i,j}$ (B) and $\chi_{j,i}$ (C) upon angular velocity during two time periods when $\dot{q}<0$ and $\dot{q}>0$ (gray shading; see main text for further explanation).

Step 1: Deriving chi rates from spike train pairs. The firing rate of a single spike train is inversely proportional to the interspike intervals (ISIs) between pairs of spikes in the train, and can thus be measured via leaky integration of the spike train (Eq. 5). We analogously define the *chi rate* for a pair of spike trains, *i* and *j*, to be a quantity that is inversely proportional to ISIs between pairs of spikes where one spike comes from neuron *i*, and the other from neuron *j*:

$$\chi_{i,j} = \left(\mathbf{s}_i \circ \mathbf{r}_j\right) * \mathbf{v}, \tag{Eq. 9}$$

where $\chi_{i,j}$ denotes the chi rate between spike trains *i* and *j*, ∘ denotes the element-wise product of same-length vectors, and $\mathbf{v}$ is an exponential decay kernel (exactly as in Eq. 5, but with a different decay constant, $\tau_v$). The term $\mathbf{s}_i \circ \mathbf{r}_j$ may be regarded as a "carbon copy" of $\mathbf{s}_i$ in which spikes no longer have unit amplitude, but instead have a real-valued amplitude that is scaled by $\mathbf{r}_j$. After the spike amplitudes are scaled, they are temporally integrated by the exponential decay kernel, $\mathbf{v}$. Hence, the chi rate is derived by sequential integration with two decay constants: $\tau_u$ (Eq. 5) and $\tau_v$ (Eq. 9). For simplicity, we set $\tau_u = \tau_v$ for all simulations presented here. However, setting these decay constants to different values may sometimes help to improve the accuracy of decoding information from co-firing rates. Possible biological substrates for computing the chi rate shall be considered in the Discussion.

Figure 3 illustrates how $\chi_{i,j}$ is influenced by $\mathbf{q}$ and $\dot{\mathbf{q}}$ for the case where neurons *i* and *j* are HD cells with nearby preferred directions, $\hat{q}_i = 0°$ and $\hat{q}_j = 60°$. These two HD cells only generate spikes when the head angle is in the neighborhood of both $\hat{q}_i$ and $\hat{q}_j$, so clearly, $\chi_{i,j}$ depends upon $\mathbf{q}$. However, $\chi_{i,j}$ is also sensitive to $\dot{\mathbf{q}}$. To see why this is so, consider what happens during the two shaded periods in Figure 3, when the head angle passes through $\hat{q}_i$ and $\hat{q}_j$ in the clockwise ($\dot{\mathbf{q}} < 0$) versus counterclockwise ($\dot{\mathbf{q}} > 0$) directions. During the first period where $\dot{\mathbf{q}} < 0$, we see that $\chi_{i,j}$ (Fig. 3B, blue) grows large, but $\chi_{j,i}$ (Fig. 3C, yellow) does not. This is because neuron $j$ spikes prior to neuron $i$, so that $\mathbf{r}_j$ is nonzero when $\mathbf{s}_i$ starts firing and $\mathbf{s}_i \circ \mathbf{r}_j$ grows large (Fig. 3B), but $\mathbf{r}_i$ is near zero when $\mathbf{s}_j$ starts firing so $\mathbf{s}_j \circ \mathbf{r}_i$ does not grow large (Fig. 3C). During the second shaded period where $\dot{\mathbf{q}} > 0$, exactly the opposite occurs, and therefore, $\chi_{j,i}$ grows large but $\chi_{i,j}$ does not. Hence, $\chi_{i,j}$ for neighboring HD cells depends not only upon $\mathbf{q}$, but also upon $\dot{\mathbf{q}}$. To derive a signal that is dependent purely upon $\dot{\mathbf{q}}$ and not upon $\mathbf{q}$, we may pool $\chi_{i,j}$ values across HD cell pairs that share the same angle of separation, $\Delta\hat{q}_{i,j}$.

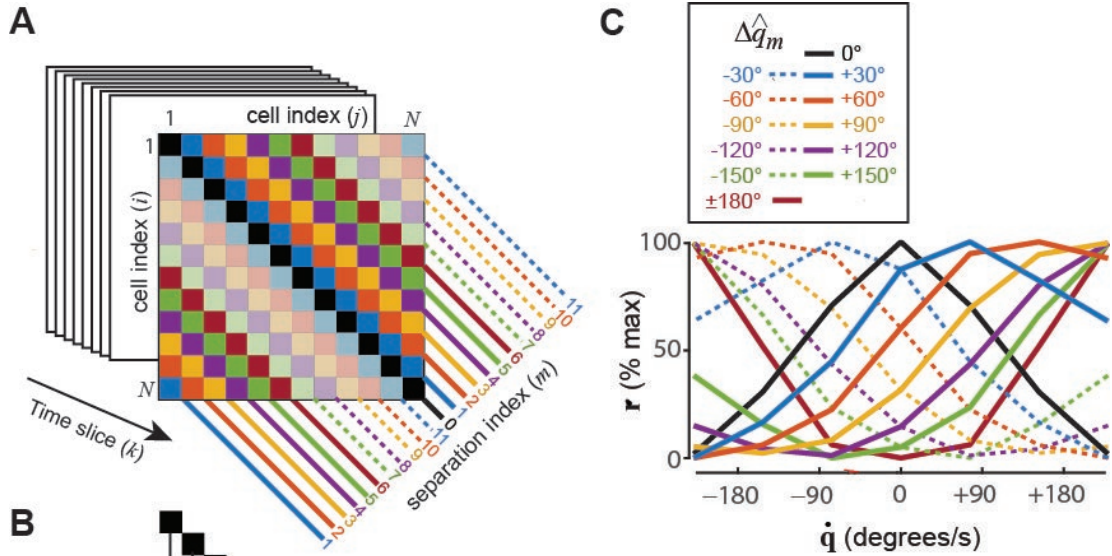

**Figure 4.** *Sigma-chi units compute co-firing rates that encode angular head velocity.* A) Chi rate matrix contains diagonal bands along which simulated HD cell pairs share the same angle of separation, $\Delta q = |q_j - q_i|$. B) Sigma-chi units compute co-firing rates by summing chi rates along each diagonal band. C) Each sigma-chi unit's co-firing rate exhibits selective tuning for angular head velocity.

Step 2: Pooling chi rates to derive co-firing rates. Given a population of $N$ spiking neurons, there are $N^2$ unique ordered pairings between the neurons, including pairings of each neuron with itself (it is necessary to consider ordered rather than unordered pairings, since $\chi_{i,j} \neq \chi_{j,i}$; as outlined in the introduction, co-firing rates are signed quantities, so $\chi_{i,j}$ and $\chi_{j,i}$ may be regarded as separate channels for conveying positive versus negative values of the co-firing rate between neurons $i$ and $j$). This forms an $N \times N$ matrix of chi rates. For HD cells, the rows and columns of this matrix can be sorted by preferred directions, $\hat{q}_1, \hat{q}_2 \cdots, \hat{q}_N$. If a time series of chi rates, $\chi_{i,j}$, is placed in each matrix entry, then the combined entries form a time series of 2D matrices, henceforth denoted as $\vec{\mathbf{X}}$ (Fig. 4A). At each time slice $t_k$, $\vec{\mathbf{X}}$ contains a 2D matrix, $\mathbf{X}_k$, in which each entry is the instantaneous chi rate, $\chi_{i,j}[t_k]$, between a pair of HD cells, $i$ and $j$. The main diagonal of $\mathbf{X}_k$ contains chi rates of HD cells paired with themselves; flanking the main diagonal there are $N-1$ diagonal bands composed from entries containing chi rates between pairs of HD cells that share the same angle of separation, $\Delta\hat{q}_{i,j} = \hat{q}_j - \hat{q}_i$, between their preferred directions. These diagonal bands wrap around at the matrix edges, since $\Delta\hat{q}_{i,j}$ is a circular variable. We shall index these diagonal bands by $m = 0,1,2, \ldots, N-1$, such that the main diagonal ($i = j$) is indexed by $m = 0$, and each non-zero index $m$ references a band containing chi rates between pairs of HD cells whose preferred directions are separated by a common angle, $\Delta\hat{q}_m = m \times 360°/N$. The activation of sigma-chi unit $m$ can then be expressed as a co-firing rate, which is the sum of all chi rates along a single band (Fig. 4B):

$$\dot{r}_m[t_k] = \sum_{i=1}^{N} \chi_{i,\text{mod}(i+m,N)}[t_k], \tag{Eq. 10}$$

where $i$ indexes each of the matrix elements along diagonal $m$. By summing up chi rates in this way, sigma-chi units acquire direction-independent sensitivity to angular velocity.

The sensitivity of co-firing rates to angular velocity depends upon the time constants of temporal integration. $\tau_u$ and $\tau_v$, and also upon the spatial separation, $\Delta\hat{q}_m$, between the preferred directions of HD cell pairs along the diagonal indexed by $m$. To see why this is so, it is helpful to think of $\dot{r}_m$ as a measurement of the change in head angle, $\Delta q$, over a specific time span, $\Delta t$. The duration of $\Delta t$ is determined by $\tau_u$ and $\tau_v$, so these decay constants should be chosen to maximize sensitivity within a representative range of angular velocities that occurs in the behavior data. In simulations presented here, we varied $\tau_u$ and $\tau_v$ to study their influence upon decoding accuracy (Fig. 2), but we did not permit different sigma-chi neurons to have different time constants (hence, in any given simulation, all of the sigma chi neurons integrated $\Delta q$ across the same span of time). However, since each sigma-chi unit has its own value of $\Delta\hat{q}_m$, different units measure $\Delta q$ with a different sized "yardstick." This endows each unit with its own preferred frequency of head rotation. The angular separations $\Delta\hat{q}_0, \Delta\hat{q}_1 \cdots, \Delta\hat{q}_{N-1}$ thus act like a set of spatial frequencies for representing angular velocity in the Fourier domain. This co-firing rate code for angular head velocity is conjugate and orthogonal to the firing rate code for angular head position. Sigma-chi units indexed by $m = 1,2, \ldots, (N/2) - 1$ have positively sloped tuning functions (Fig. 4C, solid lines), and thus generate co-firing rates that are analogous to positive frequency components. Sigma-chi units indexed by $m = (N/2) + \{1,2, \ldots, (N/2) - 1\}$ have negatively sloped tuning functions (Fig. 4C, dashed lines), and thus generate co-firing rates that are analogous to negative frequency components. The unit indexed by $m = 0$ sums over the chi rates of each HD cell with itself, and symmetrically decreases its activity for turning in either direction. The unit indexed by $m = N/2$ (assuming even-valued $N$) sums over chi rates of HD cells that have $180°$ opposing preferred directions, and symmetrically increases its activity for turning in either direction. Fig. 4C shows that activation of sigma-chi neurons is modulated by head turning in a manner that is quite similar to "angular velocity cells" that have been reported in the head direction system (Sharp, 1996; Stackman & Taube, 1998).

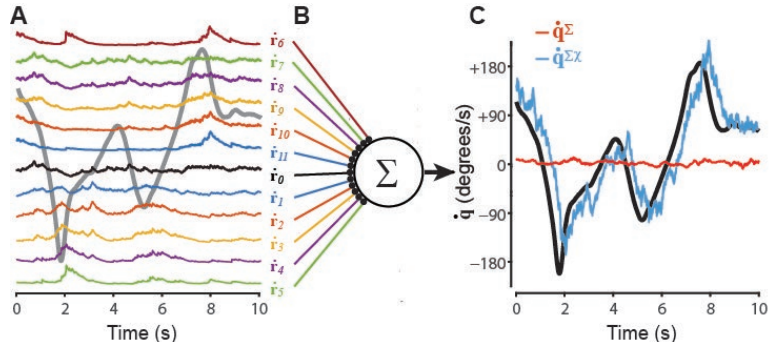

**Figure 5.** *Sigma-chi decoder recovers angular velocity from HD cell co-firing rates*. A) Colored lines show 10 s sample of co-firing rates computed by 12 sigma-chi units; gray line shows angular velocity during the simulation. B) True angular head velocity data (black) plotted alongside angular velocity signals recovered from co-firing rates via the sigma-chi decoder ($\dot{\mathbf{q}}^{\Sigma\chi}$), and from firing rates via the sigma decoder ($\dot{\mathbf{q}}^{\Sigma}$).

<u>Step 3: Decoding velocity from sigma-chi units</u>. Across time, the sigma-chi rates defined by Eq. 10 form a *co-firing rate response matrix*:

$$\dot{\mathbf{R}} = \begin{bmatrix} \dot{\mathbf{r}}_0 \\ \dot{\mathbf{r}}_1 \\ \vdots \\ \dot{\mathbf{r}}_{N-1} \end{bmatrix} = \{\dot{R}[t_0], \dot{R}[t_1], \dots, \dot{R}[t_{L-1}]\}. \tag{Eq. 11}$$

Each row of this matrix, $\dot{\mathbf{r}}_m = \{\dot{r}_m[t_0], \dot{r}_m[t_1], \cdots, \dot{r}_m[t_K]\}$, stores a time series of co-firing rates generated by sigma-chi unit $m$ (Fig. 5A). Each column, $\dot{R}[t_k] = \{\dot{r}_0[t_k], \dot{r}_1[t_k], \dots, \dot{r}_{N-1}[t_k]\}^T$, stores the instantaneous population vector of co-firing rates at time step $k$. A mapping from co-firing rates onto an estimate of the angular velocity, $\dot{\mathbf{R}} \to \dot{\mathbf{q}}^{\Sigma\chi}$, can be accomplished by computing a weighted sum of co-firing rates (Fig. 5B):

$$\dot{q}^{\Sigma\chi}[t_k] = \sum_{m=0}^{N-1} \dot{w}_m \dot{r}_m[t_k] \tag{Eq. 12}$$

where $\dot{w}_n$ is a synaptic weight assigned to the $n^{th}$ co-firing rate. To derive the synaptic weights in Eq. 12, the pseudoinverse method was used to find a weight vector that minimized the error between $\dot{\mathbf{q}}$ and $\dot{\mathbf{q}}^{\Sigma\chi}$ for a training set of HD cell spike trains. Decoding accuracy was then tested on an independent test set of spike trains, derived from different behavior data. Fig. 5C shows $\dot{\mathbf{q}}$ and $\dot{\mathbf{q}}^{\Sigma\chi}$ for a 10s example simulation where $\tau_u = \tau_v = 200$ ms. Weight fitting was repeated at different time offsets between $\dot{\mathbf{q}}$ and $\dot{\mathbf{q}}^{\Sigma\chi}$ (since a time delay is introduced by the decay kernels in the model), and the best fits were obtained when $\dot{\mathbf{q}}^{\Sigma\chi}$ was delayed from $\dot{\mathbf{q}}$ by approximately twice the value of the integration time constant (Fig. 2B). This is to be expected, since $\dot{\mathbf{R}}$ is obtained from two sequential integration steps (using decay constants $\tau_u$ and $\tau_v$), and illustrates an important asymmetry that we shall return to in the discussion: information coded in the co-firing rate channel is generally time delayed with respect to information coded in the firing rate channel.

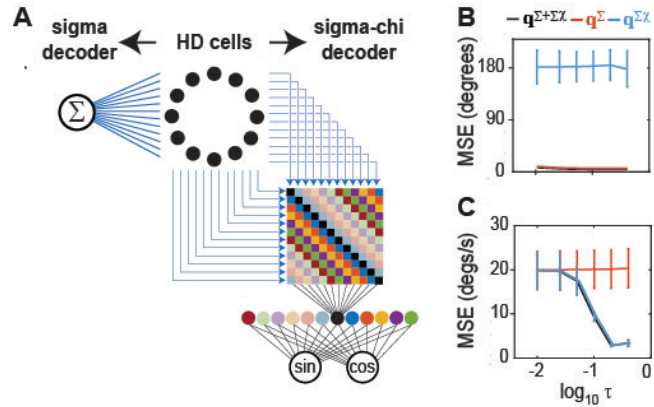

**Figure 6.** *Conjugate representations of position and velocity*. A) Spike trains from a ring of 12 simulated HD cells are fed to a sigma or sigma-chi decoder. B) Decoding error (MSE, y-axis) was computed over a range of integration constants: $\tau$ =.4, .2, .1, .05, .025, .01 s (x-axis, plotted on a $\log_{10}$ scale); head angle was accurately decoded from HD cell firing rates ($\mathbf{q}^\Sigma$) for all $\tau$, but was near chance levels when decoded from co-firing rates ($\mathbf{q}^{\Sigma\chi}$), and decoding from both channels ($\mathbf{q}^{\Sigma+\Sigma\chi}$) did not improve upon decoding from firing rates alone. C) Angular velocity was accurately decoded from simulated HD cell co-firing rates ($\dot{\mathbf{q}}^{\Sigma\chi}$) using sufficiently large integration time constant ($\tau$ >.1), but could not be decoded from firing rates ($\dot{\mathbf{q}}^\Sigma$) at any $\tau$; decoding from both channels ($\dot{\mathbf{q}}^{\Sigma+\Sigma\chi}$) did not improve upon decoding from co-firing rates alone.

Sigma-chi units perform an operation of *neural differentiation* to derive $\dot{\mathbf{q}}$ from $\mathbf{q}$. This differentiation process may be regarded as an inversion of the standard *path integration* process by which HD cells are proposed to derive $\mathbf{q}$ from $\dot{\mathbf{q}}$ in attractor-integrator network models (Zhang, 1996). Since the velocity signal encoded by $\dot{\mathbf{R}}$ is delayed in time, it shall be referred to as a *post-positional* velocity signal, to distinguish it from *pre-positional* velocity signals (such as vestibular outputs) that normally would be used for angular path integration.

*Conjugacy of firing rates and co-firing rates*

We have seen that $\mathbf{q}$ can be recovered from HD cell firing rates using the sigma decoder (Figs. 1, 2A), and $\dot{\mathbf{q}}$ can be recovered from HD cell co-firing rates using the sigma-chi decoder (Figs. 2B, 5B). What happens if we swap the decoders, and thereby attempt to recover $\mathbf{q}$ from co-firing rates with the sigma-chi decoder, and recover $\dot{\mathbf{q}}$ from firing rates with the sigma decoder? We shall write $\mathbf{q}^{\Sigma\chi}$ to denote position recovered from co-firing rates by the sigma-chi decoder, and $\dot{\mathbf{q}}^\Sigma$ to denote velocity recovered from firing rates via the sigma decoder.

To derive $\mathbf{q}^{\Sigma\chi}$, HD cell co-firing rates $\dot{\mathbf{R}}$ (instead of firing rates) were delivered as input to the sine and cosine decoder neurons (Fig. 6A). The pseudoinverse method was then used to find a weight vector that minimized error between $\mathbf{q}^{\Sigma\chi}$ and $\mathbf{q}$. Decoding accuracy was tested on a novel set of spike trains derived from head angle data that was independent from the data used to derive the weights. Fig. 1E shows that $\mathbf{q}^{\Sigma\chi}$ (blue line) generated chance-level estimates the true head angle for the 10 s example dataset. To further test whether co-firing rates contained information about $\mathbf{q}$, we used the pseudoinverse method to fit a weighted sum of both $\mathbf{R}$ and $\dot{\mathbf{R}}$ to $\mathbf{q}$, thereby obtaining a prediction $\mathbf{q}^{\Sigma+\Sigma\chi}$ of the head angle which was derived simultaneously from both firing rates and co-firing rates. Over 10 independent simulations with different head angle data (Fig. 6B), we found that $\mathbf{q}^{\Sigma}$ accurately predicted $\mathbf{q}$ regardless of the integration time constant value, but $\mathbf{q}^{\Sigma\chi}$ could not accurately predict $\mathbf{q}$ at any value of the time constant, and $\mathbf{q}^{\Sigma+\Sigma\chi}$ was never a more accurate predictor of $\mathbf{q}$ than $\mathbf{q}^{\Sigma}$. Hence, decoding head angle from firing rates and co-firing rates together was not more accurate than decoding from firing rates alone. From this, we conclude that co-firing rates of simulated HD cell spike trains only conveyed information about $\dot{\mathbf{q}}$, but not $\mathbf{q}$.

To derive $\dot{\mathbf{q}}^{\Sigma}$, the HD cell firing rates $\mathbf{R}$ (instead of co-firing rates) were delivered as input to a sigma neuron (Fig. 6A), which computed their weighted sum (Eq. 7). The pseudoinverse method was used to find a weight vector that minimized error between $\dot{\mathbf{q}}^{\Sigma}$ and $\dot{\mathbf{q}}$. Decoding accuracy was tested on a novel set of spike trains derived from independent head angle data. Fig. 5C shows that the angular velocity recovered from the sigma decoder, $\dot{\mathbf{q}}^{\Sigma}$ (red line), did not accurately estimate the true angular velocity for the 10 s example dataset. To further test whether firing rates contained information about $\dot{\mathbf{q}}$, we fit a weighted sum of both $\mathbf{R}$ and $\dot{\mathbf{R}}$ to $\dot{\mathbf{q}}$, thereby obtaining a prediction $\dot{\mathbf{q}}^{\Sigma+\Sigma\chi}$ of the angular velocity which was obtained simultaneously from both firing rates and co-firing rates. Over 10 independent simulations with different head angle data (Fig. 6C), we found that $\dot{\mathbf{q}}^{\Sigma\chi}$ accurately predicted $\dot{\mathbf{q}}$ for sufficiently large values of the integration time constant, but $\dot{\mathbf{q}}^{\Sigma}$ could not accurately predict $\dot{\mathbf{q}}$ at any value of the integration time constant, and $\dot{\mathbf{q}}^{\Sigma+\Sigma\chi}$ was never a more accurate predictor of $\dot{\mathbf{q}}$ than $\dot{\mathbf{q}}^{\Sigma\chi}$. Hence, decoding angular velocity from firing rates and co-firing rates together was not more accurate than decoding from co-firing rates alone (Fig. 5C). From this, we conclude that co-firing rates of simulated HD cell spike trains only conveyed information about $\dot{\mathbf{q}}$, but not about $\mathbf{q}$.

These simulations show that when simulated HD cell spike trains encode only the head angle (but not angular velocity) in their firing rates, the firing rate and co-firing rate channels convey orthogonal representations of position and velocity. Firing rates exclusively encode information about angular position (and not angular velocity), whereas co-firing rates exclusively encode information about angular velocity (and not angular position).

## Computing speed by differentiating grid cell spike trains

In much the same way that HD cells encode a periodic representation of an animal's angular head position, grid cells encode a periodic representation of an animal's translational position within a spatial environment. In this section, it is shown that when grid cell spike trains are fed as input to the sigma-chi decoder, sigma-chi neurons behave similarly to "speed cells" that have been reported in the hippocampus and entorhinal cortex (Kropf et al., 2015; Hinman et al., 2016; Gois & Tort, 2018).

### Simulation of grid cell spike trains

When a rat runs on a circular track, grid cells fire periodically as a function of the distance travelled around the track's circumference (Pierre-Yves et al., 2019). To simulate this periodic firing, grid cell spike trains were generated from position data that was obtained as a rat ran laps on a 1.5 m diameter circular track while harnessed to a boom arm (Jayakumar et al., 2019). The rat's positon on the track was sampled at 1 KHz by the angle of the boom arm, yielding a time series of angles measured in radians, $\boldsymbol{\phi} = \{\phi_0, \phi_1, \cdots, \phi_k, \cdots, \phi_K\}$. For simplicity, the vertex spacing of simulated grid cells was assigned to be exactly one third of the distance around the track, $\Lambda = C/3$, where $C = 471$ cm is the track circumference. Each grid cell thus fired at three stable preferred locations during every lap around the track (Fig. 7A). Grid cell tuning curves were simulated using the same Von Mises distribution described above for simulating HD cell tuning curves (Eq. 3), except that $q_k = \phi_k/3$ now represents the rat's angular position within the grid spacing interval, $\Lambda$, rather than the head azimuth angle. For grid cell simulations, Von Mises tuning parameters were $R_{max} = 100$ Hz and $\kappa = 0.25$; the spatial phases of the grids, $\{\hat{q}_1, \hat{q}_2, \cdots, \hat{q}_N\}$, were evenly spaced over $\Lambda$. As described above for HD cell simulations, Eq. 3 was used to threshold a pseudorandom number generator and stochastically determine whether grid cell $n$ fired a spike at time $t_k$. Binary values generated by this stochastic process filled a spike response matrix (Eq. 2), in which all spike trains exhibited Poisson rate statistics. In rodents, grid cell firing is typically modulated by theta rhythm, so grid cells do not generate Poisson spike trains. But here, we disregard theta modulation of grid cells for simplicity.

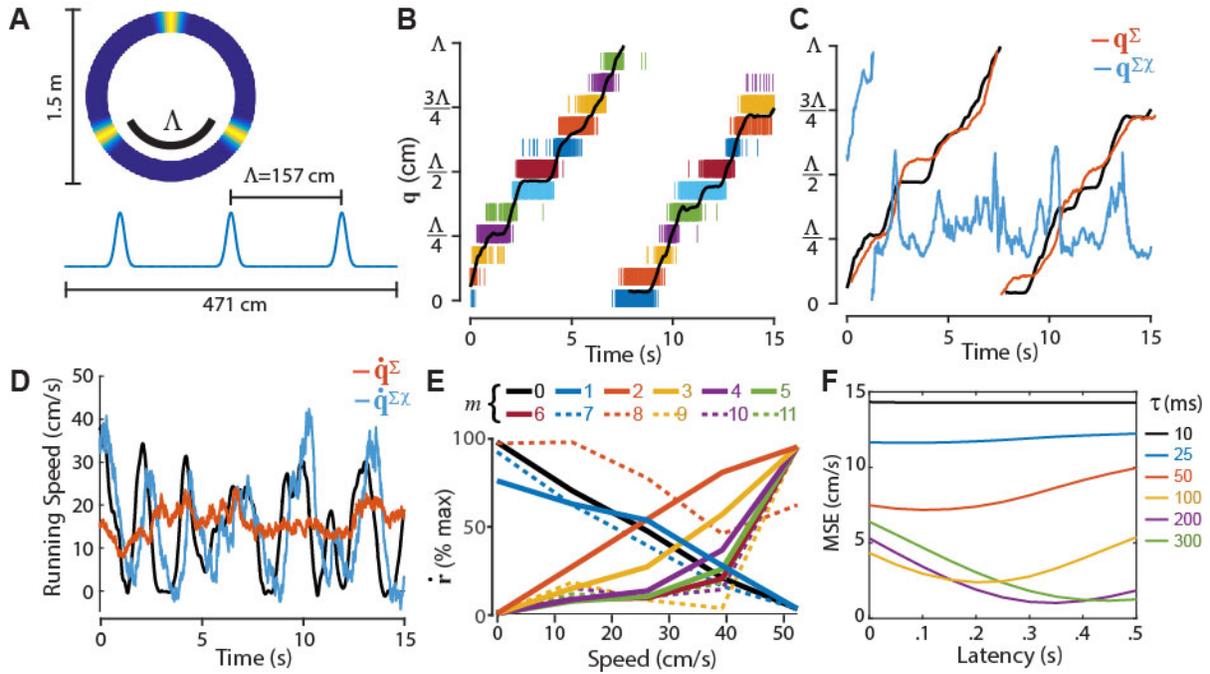

**Figure 7.** *Decoding position and running speed from grid cells.* A) Von Mises tuning functions were used to simulate periodic spatial tuning of grid cells on a circular track. B) Black line shows 15 s sample of position data, **q**; rasters show spike trains of 12 simulated grid cells. C) True position (black line) superimposed over position decoded from firing rates ($\mathbf{q}^{\Sigma}$) and co-firing rates ($\mathbf{q}^{\Sigma\chi}$). D) True running speed (black line) superimposed over running speed decoded from firing rates ($\dot{\mathbf{q}}^{\Sigma}$) and co-firing rates ($\dot{\mathbf{q}}^{\Sigma\chi}$). E) Speed tuning of co-firing rates generated by different sigma-chi neurons (indexed by *m*). F) Latency of speed tuning is proportional to the value of integration time constants.

*Decoding position and speed from grid cell spike trains*

Fig. 7B shows spike rasters for 12 simulated grid cells over a 15 s time period during which the rat ran almost 2/3 lap on the track (which is nearly 2 complete traversals of the spacing interval, $\Lambda$). Fig. 7C shows that the sigma decoder accurately recovers $\mathbf{q}$ (but fails to recover running speed) from grid cell firing rates ($\tau_u = \tau_v = 200$ ms in these simulations). Conversely, Fig. 7D shows that the sigma-chi decoder accurately recovers running speed (but fails to recover $\mathbf{q}$) from grid cell co-firing rates. When grid cell spike trains are used as inputs to the sigma-chi decoder, the sigma-chi units exhibit tuning for running speed (Fig. 7E) that is qualitatively similar to the tuning of "speed cells" in the rodent entorhinal cortex and hippocampus (Kropf et al., 2015; Hinman et al., 2016; Gois & Tort, 2018). The speed signal encoded by co-firing rates is delayed in time by an amount proportional to the integration time constants $\tau_u$ and $\tau_v$. Hence, the sigma-chi units are best tuned for the animal's past running speed (Fig. 7F), rather than present or future running speed. Sigma-chi units thus behave similarly to a subset of speed cells in the rodent brain which have been observed to lag the animal's running speed in time (Kropf et al., 2015).

**Computing positon by integrating theta cell spike trains**

Simulations presented above show that if a population of spiking neurons (HD cells or grid cells) encodes a position signal in their firing rates, then they can simultaneously encode a velocity signal in their co-firing rates. Suppose now that we wish to invert this coding scheme, so that velocity is encoded by firing rates, and position is encoded by co-firing rates. How could this be done? This may seem like a counterintuitive exercise for those accustomed to thinking about population vector codes composed from neural firing rates, because we seek to construct a neural code for position in which individual neurons exhibit no tuning whatsoever of their firing rates for position. As it turns out, there exists a class of spatial coding models called "oscillatory interference models" that are based upon precisely this idea (Burgess et al., 2007; Giocomo et al., 2007a; Blair et al., 2008). In such models, oscillatory modulation of spike trains occurs in such a way that position information is conveyed by between-cell spike intervals (i.e., spike correlations), rather than within-cell spike intervals (i.e., firing rates). Simulations presented below show that when an oscillatory interference code is constructed from simulated theta cells, position information can be recovered from theta cell co-firing rates using a sigma-chi decoding process similar to that used above for recovering velocity information from HD cell or grid cell co-firing rates, and velocity information can be recovered from theta cell firing rates using the same sigma decoding process that was used above to recover position information from HD cell or grid cell firing rates. Hence, an oscillatory interference code for position may be viewed as the conjugate inverse of a population vector code for position, because information about position and velocity is perfectly swapped out between the firing rate and co-firing rate channels.

*Simulation of theta cell spike trains*

To simulate phase coding of position on a circular track by theta cells, we assume the existence of a *reference oscillator* against which theta cells shift their phases as a function of the animal's position. To mimic the rhythmicity of theta cell spike trains, all simulations used a reference oscillator with a constant frequency of 7 Hz ($14\pi$ radians/sec). The time-varying phase of a theta cell can then be expressed as an offset, $\theta_n$, from this reference oscillator,

$$\theta_n[t_k] = \psi_n[t_k] - \psi_0[t_k], \tag{Eq. 13}$$

where $\psi_0$ is the reference phase and $\psi_n$ is the instantaneous phase of theta cell $n$. Here, all of the simulated theta cells shifted their phases against the reference oscillator (and none remained in fixed synchrony with it), so the reference oscillator was not explicitly represented by any of the theta cell spike trains. We may now define a *phase offset matrix*,

$$\mathbf{\Theta} = \begin{bmatrix} \mathbf{\theta}_1 \\ \mathbf{\theta}_2 \\ \vdots \\ \mathbf{\theta}_N \end{bmatrix} = \{\Theta_0, \Theta_1, \dots, \Theta_k, \dots, \Theta_K\}, \tag{Eq. 14}$$

where each column, $\Theta_k = \{\theta_1[t_k], \theta_2[t_k], \dots, \theta_N[t_k]\}^T$, gives the instantaneous phase offsets for $N$ theta cells at time $t_k$, and each row, $\mathbf{\theta}_n = \{\theta_n[t_0], \theta_n[t_1], \cdots, \theta_n[t_K]\}$, is a time series of phase offsets for theta cell $n$. Since $\theta_n$ is an angle, it can encode the animal's position on a spatially periodic interval. The animal's position within this interval shall be denoted $q_n$, defined as follows:

$$q_n[t_k] = 2\pi \frac{x[t_k]}{\lambda_n}, \tag{Eq. 15}$$

where $x[t_k]$ is the total integrated distance that the animal has travelled around the track at time $t_k$, and $\lambda_n$ is the distance over which $\theta_n$ shifts by one full theta cycle against the reference oscillator. The dependence of $\theta_n$ upon angular position may then be written

$$\theta_n[t_k] = \hat{\theta}_n + q_n[t_k], \tag{Eq. 16}$$

where $\hat{\theta}_n$ is a phase offset parameter for theta cell $n$. Eq. 16 forces theta cells to behave as velocity controlled oscillators (VCOs) like those used in oscillatory interference models of spatial coding (Burgess et al., 2007). As in these prior models, the burst frequency, $\omega_n$, of each theta cell varies as a function of the animal's running speed,

$$\omega_n[t_k] = \omega_0 + 2\pi \frac{v[t_k]}{\lambda_n}, \tag{Eq. 17}$$

where $v[t_k]$ is the running speed at time $t_k$, and $\omega_0$ is the frequency of the reference oscillator in radians/s. In simulations presented below, we shall permit $\lambda_n$ to take either positive or negative values. If $\lambda_n > 0$, then $\omega_n$ increases with running speed, and $\theta_n$ precesses in phase against the reference oscillator. By contrast, if $\lambda_n < 0$, then $\omega_n$ decreases with running speed, and the reference oscillator precesses in phase against $\theta_n$.

   To compute the probability of spiking at each time step, we first created a "seed spike train" for each theta cell, denoted $\tilde{\mathbf{s}}_n$. This is a binary spike response function (Eq. 1) containing a single spike at every time step where theta cell $n$ passes through perfect phase synchrony with the reference oscillator. The probability of spiking for theta cell $n$ at each time step was then computed by convolving the seed spike train with a Gaussian kernel:

$$p_n[t_k] = R_{max}(\tilde{\mathbf{s}}_n * \mathbf{g}), \tag{Eq. 18}$$

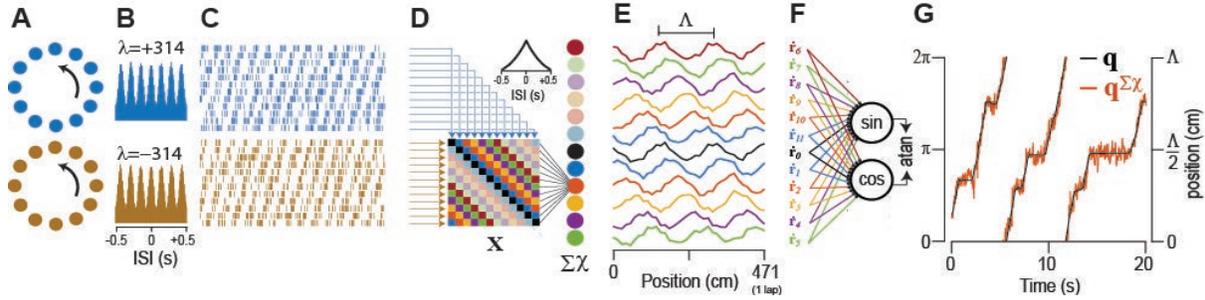

**Figure 8.** *Sigma-chi decoder recovers position from theta cell co-firing rates*. A) Two sets of 12 simulated theta cells resided in ring oscillators with phase slopes of equal and opposite sign. B) Autocorrelograms for theta cells in each ring. C) Rasters show 2 s sample of theta-modulated spike trains in the two rings. D) Sigma-chi decoder converts simulated theta cell spike trains into chi rates (**X**) and then into co-firing rates encoded by activity of sigma-chi neurons (inset shows autocorrelogram for a sigma-chi neuron). E) Sigma-chi neurons behave like grid cells with periodic spatial tuning on the circular track. F) Sine and cosine components of angular position on the interval $\Lambda$ are decoded via vector summation. G) The atan2 function converts output from sine and cosine decoder neurons into an accurate prediction, $\mathbf{q}^{\Sigma\chi}$), of position on the track.

where $R_{max}$ sets the maximum spike probability within a single time step, and **g** is a Gaussian kernel of unit amplitude. Here we used $\sigma = 25$ ms as the width of the Gaussian kernel in all simulations. To prevent the spike probability from exceeding $R_{max}$ in any time bin, convolution was performed separately on even and odd numbered spikes within $\tilde{\mathbf{s}}_n$ (see Methods).

*Ring oscillators*

To construct a phase code for position, we shall group theta cells into subpopulations that form *ring oscillators* (Blair et al., 2014). Each ring oscillator is a circular array of theta cells through which a "bump" of activity circulates at a rate of once per theta cycle. To simulate bump circulation, phase offset parameters of theta cells within a ring are staggered at even spacings throughout the cycle by assigning $\hat{\theta}_p = 2\pi(p-1)/P$, where $p$ indexes theta cells within the ring, and $P$ is the total number of theta cells within a ring ($P = 12$ for all simulations presented here). Hence, theta cells exhibit preferred phases that are evenly space around the theta cycle, in much the same way that the preferred directions of HD cells were evenly spaced around the circle in simulations above.

Our present objective shall be to construct a phase code that perfectly segregates information about position and velocity into the co-firing rate and firing rate channels, respectively. That is, the co-firing rate channel shall only convey information about position (but not velocity), and the firing rate channel shall only convey information about velocity (but not position). To achieve this perfect segregation, it is necessary to define pairs of ring oscillators, $a$ and $b$, composed from theta cells with phase slopes of the same magnitude but opposing sign (Fig. 8A). Hence, if the phase slope for theta cells in ring $a$ is denoted by $\lambda_a$, then the slope for cells in ring $b$ must be $\lambda_b = -\lambda_a$. Assuming $\lambda_a > 0$, then theta cells in ring $a$ precess (shift backward in phase) against the reference oscillator by one cycle per traversal of the distance $\lambda_a$, whereas cells in ring $b$ process (shift forward in phase) against the reference oscillator by one cycle over the same distance. Fig. 8B shows example autocorrelograms for theta cells residing in a pair of ring oscillators with $\lambda_a = C/1.5 = 314$ cm and $\lambda_a = -C/1.5 = -314$ cm. Raster plots of spike trains for these theta cells are shown in Fig. 8C.

### Decoding position from theta cell co-firing rates

Since theta cell phases are dependent upon $q_n$ (Eq. 16), the rat's angular position on the track can be decoded from co-firing rates via the sigma-chi decoding process, which in this case has four steps: $\mathbf{S} \to \vec{\mathbf{X}} \to \dot{\mathbf{R}} \to \mathbf{Y} \to \mathbf{q}^{\Sigma\chi}$. In this section, we shall write $\mathbf{q}^{\Sigma\chi}$ to denote the time series of angular track positions recovered from co-firing rates via the sigma-chi decoder, and $\mathbf{q}$ to denote the time series of true positions.

Step 1: Deriving chi rates from spike trains. The first decoding step, $\mathbf{S} \to \vec{\mathbf{X}}$, converts pairs of theta cell spike trains into chi rates, $\boldsymbol{\chi}_{i,j}$, using Eq. 9 above. To recover position information from $\boldsymbol{\chi}_{i,j}$, it is necessary for theta cells $i$ and $j$ to reside in different ring oscillators, so that the two cells will shift phase against one another as a function of the animal's position. Moreover, to obtain co-firing rates that exclusively encode position and not velocity (which is the goal of our current didactic exercise, but is by no means a requirement for neural coding of position and velocity in the brain), it is necessary for the rings containing cells $i$ and $j$ to have phase slopes of identical magnitude but opposing sign (that is, $\lambda_i = -\lambda_j$). A pair of theta cells meeting these requirements will shift through one cycle of phase against each other each time the animal traverses a distance equal to $\Lambda = |\lambda|/2$, where $|\lambda|$ is the shared magnitude of the phase slope for cells $i$ and $j$. Fig. 8 shows an example with $|\lambda| = 314$ cm, so that $\Lambda = 157$ cm.

Step 2: Pooling chi rates to derive co-firing rates. The second decoding step, $\vec{\mathbf{X}} \to \dot{\mathbf{R}}$, pools chi rates from multiple theta cell pairs to derive a vector of co-firing rates. Given two ring oscillators that each contain $P$ theta cells, there are $P^2$ unique ordered pairings between theta cells that reside in different rings, forming a $P \times P$ matrix of chi values (Fig. 8D). The rows and columns of this matrix can be sorted by the preferred phases, $\hat{\theta}_1, \hat{\theta}_2 \cdots, \hat{\theta}_P$, of theta cells in each ring. At each time step, $t_k$, we may define a 2D matrix, $\mathbf{X}_k$, in which each entry contains the instantaneous chi rate, $\chi_{i,j}[t_k]$, for theta cells $i$ and $j$. This is exactly the same matrix structure that was defined for pairs of HD cells in Fig. 4A above, except that here, the preferred firing phase of theta cells is substituted for the preferred head angle of HD cells. Hence, the main diagonal of $\mathbf{X}_k$ contains chi rates for pairs of theta cells that share the same phase offset parameter, $\hat{\theta}_n$, within their respective ring oscillators. Flanking the main diagonal, there are $N - 1$ diagonal bands composed from entries containing chi rates for pairs of theta cells that share the same difference between their phase offset parameters, $\Delta\hat{\theta}_{i,j} = \hat{\theta}_j - \hat{\theta}_i$. As illustrated above (Fig. 4A), we may index these diagonal bands by $m = 0,1,2,...,P - 1$, such that the main diagonal ($i = j$) is indexed by $m = 0$, and each non-zero index $m$ references a band containing chi rates between pairs of theta cells with preferred phases separated by a common offset, $\Delta\hat{\theta}_m = m \times 2\pi/P$. Activation of sigma-chi unit $m$ can then be expressed as a co-firing rate, $\dot{r}_m$, obtained by summing all chi rates along a single band, as in Eq. 10 above. Since sigma-chi units pool inputs across all phases of theta, the time series $\dot{\mathbf{r}}_m$ is not itself modulated by theta (Fig. 8D, inset). Fig. 8E shows that each co-firing rate $\dot{\mathbf{r}}_m$ exhibits spatially periodic positional tuning, and therefore, each sigma-chi unit behaves like a grid cell with vertex spacing equal to $\Lambda$ and spatial phase equal to $\Delta\hat{\theta}_m/2\pi$. It is thus possible to decode angular position, $\phi$, from the vector of co-firing rates, in much the same way that position was decoded above from firing rates of simulated grid cell spike trains (Fig. 7C).

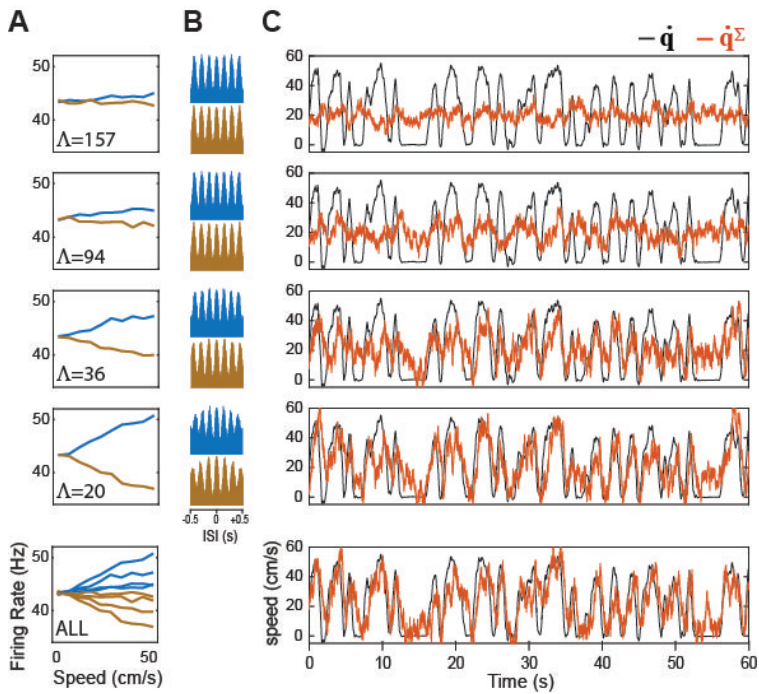

**Figure 9.** *Sigma decoder recovers velocity from theta cell firing rates.* A) Graphs plot dependence of simulated theta cell firing rates (y-axis) upon running speed (x-axis) for different values of the phase slope parameter, $\Lambda$; theta cells with positive and negative phase slopes are plotted blue and brown, respectively. B) Autocorrelograms for theta cells plotted in 'A.' C) Actual running speed $\dot{\mathbf{q}}$ (black) superimposed with decoded speed $\dot{\mathbf{q}}^{\Sigma}$ (red) during a 60 s period of running on the circular track; top four graphs show $\dot{\mathbf{q}}^{\Sigma}$ decoded from 24 theta cells with phase slopes (12 positive, 12 negative) in 'A'; bottom graph shows running speed decoded from all theta cells in top four graphs combined (96 theta cell in total, 48 with positive and 48 with negative phase slopes).

Step 3: Converting co-firing rates into angle components. The third decoding step, $\dot{\mathbf{R}} \rightarrow \mathbf{Y}$, converts co-firing rates into sine and cosine components of the animal's periodic angular position, $\phi$, within the interval $\Lambda$. A pair of decoder neurons extracts the sine and cosine components of $\phi$ (Fig. 8F). The pseudoinverse method was used to assign weights minimizing the error for estimating $\sin(\phi)$ and $\cos(\phi)$.

Step 4: Converting angle components into periodic position. The fourth decoding step, $\mathbf{Y} \rightarrow \mathbf{q}^{\Sigma\chi}$, converts the sine and cosine components of $\phi$ into a prediction of the animal's position within the interval $\Lambda$. The atan2 function was used to obtain $\phi$ in radians from $\sin(\phi)$ and $\cos(\phi)$, and radians were then converted to distance by $q = \Lambda \times \phi/2\pi$ (Fig 8G).

### Decoding velocity from theta cell firing rates

We now show that the animal's running speed can be decoded from theta cell firing rates via the sigma decoding process, which in this case has two steps: $\mathbf{S} \rightarrow \mathbf{R} \rightarrow \dot{\mathbf{q}}^{\Sigma}$. In this section, we shall write $\dot{\mathbf{q}}^{\Sigma}$ to denote the time series of running speeds recovered from the sigma decoder, and $\dot{\mathbf{q}}$ to denote the time series of true running speeds.

Step 1: Deriving theta cell firing rates from their spike trains. The first decoding step, $\mathbf{S} \rightarrow \mathbf{R}$, converts theta cell spike trains into firing rates. For simulations presented here, theta cell spike trains were converted into firing rates using Eq. 5. As explained above, each theta cell's burst frequency varies with running speed in accordance with Eq. 16. However, we define "firing rate" to mean the rate at which individual spikes occur—not the rate at which theta bursts occur—so it does not automatically follow from Eq. 17 that theta cell firing rates must vary with running speed. Nonetheless, since a theta cell's spike probability oscillates with its burst frequency (Eq. 18), it turns out that theta cells do indeed behave as "speed cells" with firing rates that are modulated by running speed. Firing rates of theta cells residing in paired rings $a$ and $b$ were modulated by running speed with a steepness that was inversely proportional to their phase slopes, $\lambda_a$ and $\lambda_b$ (Fig. 9A). Theta cells with positive phase slopes ($\lambda_a > 0$) increased their firing rates with running speed, and those with negative phase slopes ($\lambda_b < 0$) decreased their firing rates with running speed. Fig. 9B shows autocorrelograms for simulated theta cells over a range of different phase slopes.

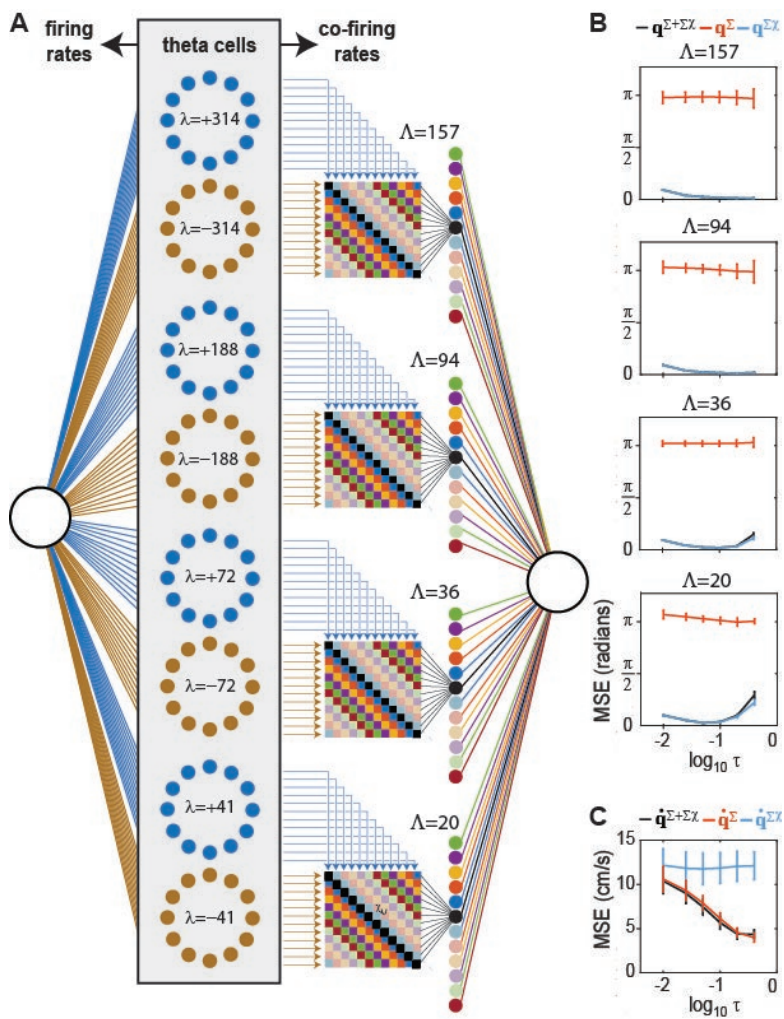

**Figure 10.** *Segregation of position and velocity into co-firing and firing rate channels.* A) Four complementary pairs of ring oscillators contain a total of 96 simulated theta cells; firing rates are fed to a sigma decoder, and co-firing rates are fed to a sigma-chi decoder. B) Angular position (**q**, y-axis) is accurately decoded from co-firing rates ($\mathbf{q}^{\Sigma\chi}$) but not firing rates ($\mathbf{q}^{\Sigma}$) of theta cells for all spatial periods (L) and integration time constants (τ =.4, .2, .1, .05, .025, or .01 s; x-axis plots t on a log scale). C) Running speed ($\dot{\mathbf{q}}$, y-axis) is accurately decoded from firing rates ($\dot{\mathbf{q}}^{\Sigma}$) but not co-firing rates ($\dot{\mathbf{q}}^{\Sigma\chi}$) of theta cells for sufficiently large integration time constants (x-axis).

Step 2: Decoding running speed from firing rates. The second decoding step, $\mathbf{R} \rightarrow \dot{\mathbf{q}}^{\Sigma}$, converts theta cell firing rates into an estimate of the animal's running speed. We have already seen that theta cell firing rates are modulated by running speed, so in principle, it should be easy to decode the animal's running speed from theta cell firing rates. However, when we used the pseudoinverse method to derive weights that minimized the error between $\dot{\mathbf{q}}^{\Sigma}$ and $\dot{\mathbf{q}}$ for theta cells with $|\lambda| = 314$ cm, the fit was not good when the decoder was tested on data independent from the training set (Fig 9C, top panel). This is because the accuracy of speed decoding is proportional to the slope with which velocity modulates theta cell firing rates, which in turn is inversely proportional to $|\lambda|$, and setting $|\lambda| = 314$ cm yields a slope that is too shallow for accurate speed decoding. This problem is easily remedied by using theta cells with steeper phase slopes, which increases the slope of velocity modulation (and also decreases the vertex spacing, $\Lambda$, of grid cells formed via the sigma-chi decoding process; see Fig. 10A). Fig. 9C shows that velocity decoding progressively improved with steeper slopes of velocity modulation, becoming quite accurate for $|\lambda| = 41$ cm, which yields a vertex spacing of $\Lambda = 20.5$ cm for grid cells simulated by sigma-chi neurons; this is within the range of experimentally observed vertex spacings for grid cells (Hafting et al., 2005).

There is no biological requirement for all ring oscillators to share the same value of $|\lambda|$; in fact, different values of $|\lambda|$ yield sigma-chi neurons that mimic grid cells with different vertex spacings. Thus, we simulated multiple ring oscillators with different phase slopes. Four pairs of ring oscillators (8 rings in all, with 12 theta cells in each ring) containing a total of 96 theta cells were assigned phase slopes $\lambda_1 = \pm 314$ cm, $\lambda_2 = \pm 188$ cm, $\lambda_3 = \pm 72$ cm, and $\lambda_4 = \pm 41$ cm. Fig. 10B shows that when spike trains from these four ring pairs were fed into a sigma-chi decoder (configured so that chi rates were derived within but not between ring pairs), it was possible to accurately recover the animal's position $\mathbf{q}$ on intervals of length $\Lambda_1 = 157$ cm, $\Lambda_2 = 94$ cm, $\Lambda_3 = 36$ cm, and $\Lambda_4 = 20.5$ cm, which are the vertex spacings of grid cells simulated by sigma-chi neurons formed from each ring pair. A sigma neuron summing inputs from all 96 theta cells (using weights assigned via the pseudoinverse method on an independent training dataset) accurately recovered the animal's running speed from theta cell firing rates (Fig. 10C).

*Conjugate coding of position and velocity*

Results shown in Figs. 8-10 demonstrate that theta cells can simultaneously encode position information in their co-firing rates, while also encoding velocity (running speed) information in their firing rates. As explained above, the theta cell population was constructed to contain pairs of ring oscillators with equal and opposite phase slopes (Fig. 9A), so that $\mathbf{q}$ was decoded from each ring pair on a spatially periodic interval of length $\Lambda$ (which was determined by the phase slopes for that ring pair). We now show that if co-firing rates are computed exclusively within (and not between) these complementary ring pairs, then position ($\mathbf{q}$) and running speed ($\dot{\mathbf{q}}$) information become perfectly segregated into the co-firing and firing rate channels, respectively. This can be demonstrated by swapping the decoders, and thereby attempting to recover $\dot{\mathbf{q}}$ from co-firing rates and $\mathbf{q}$ from firing rates. We shall write $\mathbf{q}^{\Sigma}$ to denote position recovered from firing rates, and $\dot{\mathbf{q}}^{\Sigma\chi}$ to denote velocity recovered from co-firing rates.

To derive $\dot{\mathbf{q}}^{\Sigma\chi}$, a weighted sum of theta cell co-firing rates $\{\dot{\mathbf{r}}_1, \dot{\mathbf{r}}_2, ..., \dot{\mathbf{r}}_N\}^T$ was computed, retaining the constraint that co-firing rates were only computed within (and not between) complementary ring pairs. The pseudoinverse method was then used to find a weight vector that minimized error between $\dot{\mathbf{q}}^{\Sigma\chi}$ and $\dot{\mathbf{q}}$. Decoding accuracy was tested on a novel set of spike trains derived from independent behavior data. Fig. 10C shows that $\dot{\mathbf{q}}^{\Sigma\chi}$ did not accurately estimate the true running speed. To further test whether co-firing rates contained information about $\dot{\mathbf{q}}$, we attempted to recover running speed from a weighted sum of both firing rates and co-firing rates, $\dot{\mathbf{q}}^{\Sigma+\Sigma\chi}$, and used the pseudoinverse method to minimize error between $\dot{\mathbf{q}}^{\Sigma+\Sigma\chi}$ and $\dot{\mathbf{q}}$. Over 10 independent simulations with different behavior data, we found that decoding $\dot{\mathbf{q}}$ from theta cell firing rates and co-firing rates together was not more accurate than decoding from firing rates alone (Fig. 10C). That is, the accuracy of $\dot{\mathbf{q}}^{\Sigma+\Sigma\chi}$ was not better than the accuracy of $\dot{\mathbf{q}}^{\Sigma}$. From this, we conclude that firing rates of simulated theta cells only conveyed information about $\dot{\mathbf{q}}$, and not about $\mathbf{q}$.

To derive $\mathbf{q}^{\Sigma}$, a weighted sum of theta cell firing rates $\{\mathbf{r}_1, \mathbf{r}_2, ..., \mathbf{r}_N\}^T$ was computed, and the pseudoinverse method was used to find a weight vector that minimized error between $\mathbf{q}^{\Sigma}$ and $\mathbf{q}$. Decoding accuracy was then tested on a novel set of spike trains derived from independent behavior data. Fig. 5B shows that $\mathbf{q}^{\Sigma}$ did not accurately estimate the animal's true position on the interval $\Lambda$. To further test whether firing rates contained information about $\mathbf{q}$, we attempted to recover running speed from a weighted sum of both firing rates and co-firing rates, $\mathbf{q}^{\Sigma+\Sigma\chi}$, and used the pseudoinverse method to minimize error between $\mathbf{q}^{\Sigma+\Sigma\chi}$ and $\mathbf{q}$. Over 10 independent simulations with different behavior data, we found that decoding $\mathbf{q}$ from theta cell firing rates and co-firing rates together was no more accurate than decoding from co-firing rates alone (Fig. 10B). From this, we conclude that if co-firing rates are computed from complementary pairs of theta ring oscillators, then they only convey information about $\mathbf{q}$, and not about $\dot{\mathbf{q}}$.

*Co-firing rates of non-complementary ring oscillator pairs*

Given a set of $N$ ring oscillators, there are $(N^2 - N)/2$ unique non-ordered pairings between different rings. In the simulations of Fig. 10, we used 8 ring oscillators (4 pairs of complementary rings), which yields a total of 28 non-ordered pairings. However, we only computed co-firing rates from 4 of these 28 possible pairings, because as explained above, the 4 pairings between complementary rings have the special property of yielding co-firing rates which only contain position (and not velocity) information.

It is certainly possible to derive co-firing rates from non-complementary ring pairs, but co-firing rates computed in this way will contain information about both position and velocity, rather than just position. To demonstrate this, Fig. 11 shows results from simulations of 4 ring oscillators—each containing 12 theta cells (for a total of 48 theta cells in all)—which were all assigned to have positive phase slopes: $\lambda_1 = +314$ cm, $\lambda_2 = +188$ cm, $\lambda_3 = +55$ cm, and $\lambda_4 = +39$ cm. There are 6 unique non-ordered pairings among these 4 rings, and since all rings have positive phase slopes, there are no complementary pairs. Fig. 11B shows that position can only be decoded from co-firing rates (but not firing rates) of theta cells residing in non-complementary ring pairs, a result similar to that shown above for complementary ring pairs. However, Fig. 11C shows that running speed can now be recovered from either firing rates or co-firing rates of theta cells residing in non-complementary ring pairs, and the most accurate decoding is obtained when running speed is recovered from both together. This is in marked contrast with simulations in Fig. 10, where running speed was not decodable from co-firing rates of complementary ring pairs. However, Fig. 11C shows that position can still only be decoded from co-firing rates (but not firing rates) of theta cells residing in non-complementary ring pairs, a result similar to that shown for complementary ring pairs in Fig 10. We shall now add positional modulation of firing rates to the non-complementary ring model of Fig. 11, to demonstrate how the same information can be encoded by a population of speed cells versus grid cells, despite their different firing rate tuning.

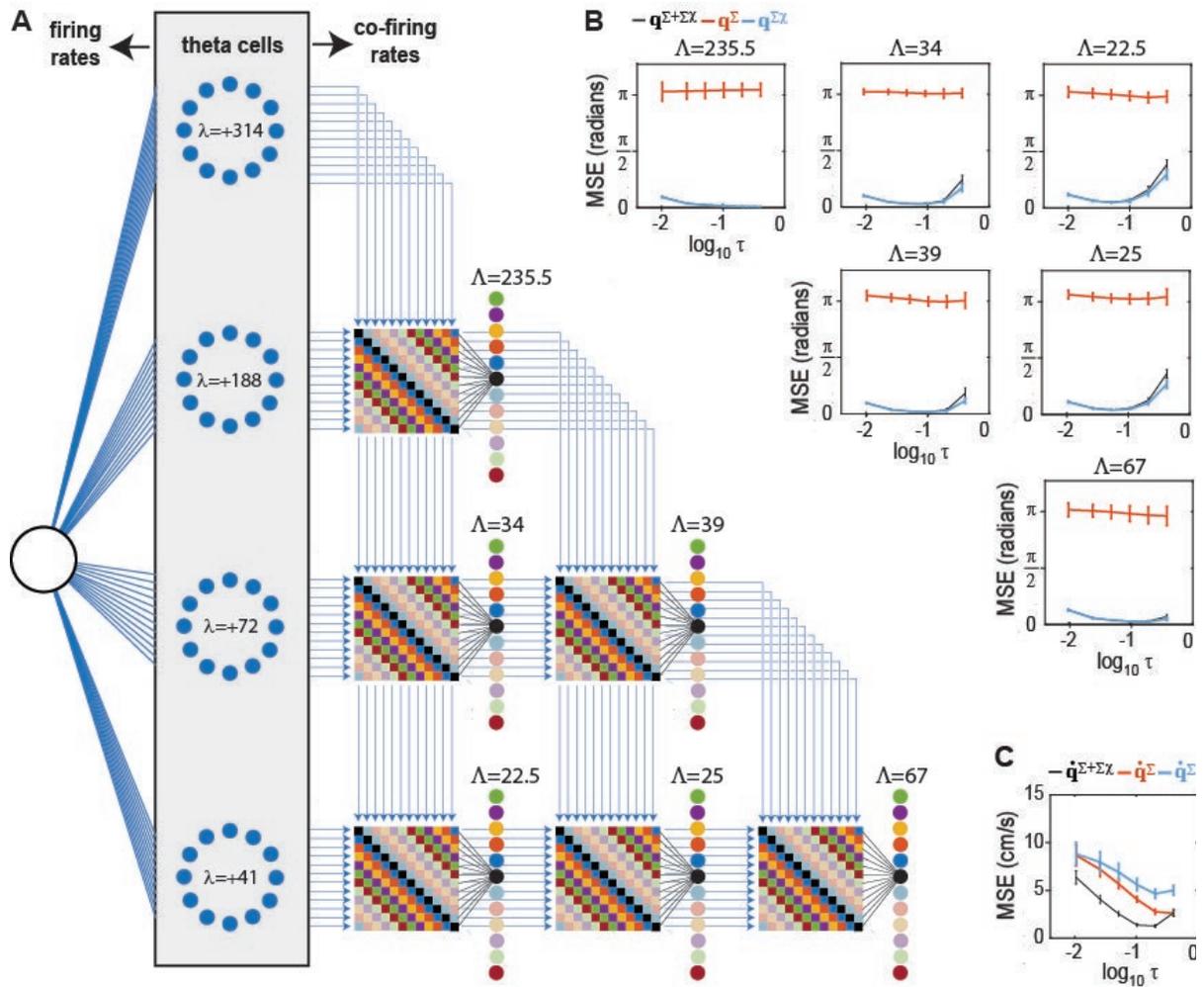

**Figure 11.** *Position and velocity coding with non-complementary ring oscillators*. A) Co-firing rates are derived from six unique pairings among four ring oscillators, all with positive phase slopes. B) Angular position can be decoded from co-firing rates on six different length intervals ($\Lambda$ =22.5, 25, 34, 39, 67, 235.5 cm); decoding error (MSE, y-axis) is low when angular position is decoded from theta cell co-firing rates ($\mathbf{q}^{\Sigma\chi}$) but near chance levels when decoded from firing rates ($\mathbf{q}^{\Sigma}$); x-axis plots integration time constant ($\tau$ =.4, .2, .1, .05, .025, .01 s) on a log10 scale. C) Running speed ($\dot{\mathbf{q}}$, y-axis) can be decoded from either firing rates ($\dot{\mathbf{q}}^{\Sigma}$) or co-firing rates ($\dot{\mathbf{q}}^{\Sigma\chi}$) of simulated theta cells, and is most accurate when decoded from both together ($\dot{\mathbf{q}}^{\Sigma+\Sigma\chi}$).

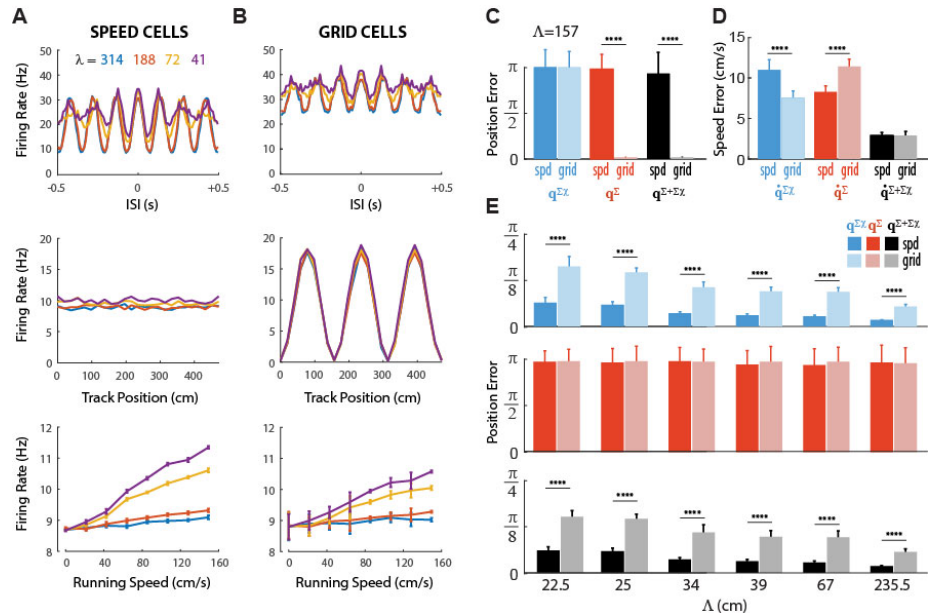

**Figure 12.** *Adding position information to firing rate channel comes at the expense of information in other channels*. A,B) Graphs show autocorrelograms (top), position tuning (middle), and speed tuning (bottom) for simulated speed cells (A) and grid cells (B). C,D) Mean position error (C) at length scale $\Lambda = 157$ cm and speed error for the sigma, sigma-chi, and combined decoders. E) Mean position error at length scales encoded by the six ring pairs for the sigma-chi (top), sigma (middle), and combined (bottom) decoders. All simulations conducted using $\tau = 100$ ms.

*Speed and grid cells can encode the same information*

    Fig. 12 compares results from simulations of speed cells versus grid cells; for simplicity, all simulations used $\tau_u = \tau_v = 100$ ms. Speed cells were simulated using the same model as in Fig. 11, as a population of theta neurons residing in four non-complementary ring oscillators. Hence, speed cells exhibited strong theta modulation of their spike trains (Fig. 12A, top), linear modulation of their firing rates by running speed (Fig. 12A, bottom), and no modulation of their firing rates by position on the track (Fig. 12A, middle). Grid cell spike trains were simulated by probabilistically re-sorting the speed cell spike trains in time (see Methods), so that their firing rates became periodically tuned for position with a period of $\Lambda = 157$ cm (spatial phases of the 12 cells in each ring were evenly spaced across $\Lambda$). Importantly, this re-sorting was done in such a way that the total number of spikes fired by each neuron was preserved, so that no position or velocity information was added or subtracted merely by increasing or decreasing the total number of spikes. But since the re-sorting was probabilistic, some (but not all) of the original theta and speed modulation in the pre-sorted spike trains was preserved after sorting. Re-sorting spike trains to convert speed cells into grid cells thus endowed the neurons with periodic spatial tuning (Fig. 12B, middle), while also reducing the depth of theta modulation (Fig. 12B, top) and the slope of speed modulation (Fig. 12B, bottom).

    We then analyzed how adding position information to the firing rate channel (by re-sorting spikes) affected the amount of information about position and velocity that was conveyed by firing rates versus co-firing rates. Fig. 12C (red bars) shows that when the rat's position

within the interval $\Lambda = 157$ cm was decoded from firing rates, the decoding error was near chance levels for speed cells, but near zero for grid cells. Hence, as expected, position information was conveyed by the firing rates of grid cells but not speed cells. Fig. 12C (blue bars) shows that when the rat's position within the interval $\Lambda = 157$ cm was decoded from co-firing rates, the decoding error was near chance levels for both speed and grid cells, because no pair of ring oscillators encoded the rat's position on a length scale of $\Lambda = 157$ cm in their co-firing rates. By contrast, when the rat's position was decoded from co-firing rates on any of the length scales that were encoded by one of the six ring oscillator pairs, the decoding error was very small for speed cells (as already shown above in Fig. 11) and significantly larger (though still far below chance levels) for grid cells (Fig. 12E, blue bars). This demonstrates that when speed cells were converted to grid cells, adding position information to the firing rate channel *came at the expense of position information in the co-firing rate channel*, in accordance with an uncertainty trade-off between firing rates and co-firing rates: the more information the firing rate channel conveys about a given variable (in this case, positon), the less information about that variable can be conveyed in the conjugate co-firing rate channel, and vice versa. Note that when position was decoded from firing rates on the length scales encoded by co-firing rates (Fig. 12E, red bars), the decoding error was near chance levels for both speed and grid cells; for this reason, the error when decoding from both firing and co-firing rates together was nearly identical to the error when decoding from co-firing rates alone (Fig. 12E, black bars).

Fig. 12D (black bars) shows that when the rat's running speed was decoded from firing rates and co-firing rates together, the decoding error was nearly identical for speed cells and grid cells. Hence, adding position information to the firing rate channel did not come at the expense of the total velocity information; together, the firing and co-firing rate channels conveyed just as much information about running speed before versus after position information was added to the firing rate channel. But importantly, *information about running speed was re-distributed between the firing rate and co-firing rate channels*. Fig. 12D shows that speed cells had lower sigma than sigma-chi decoding error for running speed, and thus conveyed more information about velocity in their firing rates than co-firing rates; since speed cells were more strongly modulated by theta than grid cells, this is fully consistent with simulations from Fig. 9 above, showing that theta cells preferentially encode velocity information in their firing rates. By contrast, grid cells had lower sigma-chi than sigma decoding error for running speed, and thus conveyed more information about velocity in their co-firing rates than firing rates; this is fully consistent with simulations in Fig. 5-7 above, showing that neurons with position-tuned firing rates (such as HD cells and grid cells) preferentially encode velocity information in their co-firing rates. But importantly, the total amount of velocity information conveyed by firing rates and co-firing rates together was not different for speed and grid cells. What this suggests is that position and velocity information *only compete with one another for coding capacity within a single coding channel (firing rates or co-firing rates), but do not compete with one another for capacity across coding channels.* We take this as evidence for an uncertainty principle that governs neural coding of conjugate variables by conjugate coding channels. We shall now articulate this hypothesized uncertainty principle in a formal and falsifiable way.

**An uncertainty principle for neural coding of conjugate variables**

In quantum physics, the uncertainty principle states that the more information we have about a particle's position, the less we can know about its momentum (or velocity), and vice versa. A standard mathematical formulation of the uncertainty principle is written $\sigma_x \sigma_p \geq \hbar/2$ (where $\hbar$ is the reduced Planck constant), placing a lower bound on the product of variances for the probability distributions of position, $\sigma_x$, and momentum, $\sigma_p$ (Kennard, 1927). Based on the results of simulations above, we shall now formally hypothesize (but not conclusively prove) a similar set of uncertainty principles for neural coding of conjugate variables. We shall do this by stating these principles using mathematical expressions that have a form similar to the uncertainty principle from physics.

*Between-channel uncertainty trade-off*

Assume that a population of neurons generates a set of spike trains **S,** containing some fixed number of spikes $\Sigma S$. Let us first consider the case where the population encodes information about a single unidimensional variable, $q$. As we have seen, information about $q$ can be encoded by two channels embedded in **S**: firing rates $(R)$ and co-firing rates $(\dot{R})$. Hence, **S** simultaneously implements two neural codes, $q \to R$ and $q \to \dot{R}$. We posit that these two codes are conjugates of one another, and thus, there is an uncertainty trade-off between them: the more information one code conveys about $q$, the less information the other code can convey about $q$. Each channel thus has its own limited capacity to convey information about $q$, which is directly proportional to $\Sigma S$ and inversely proportional to the amount of information that the other channel conveys about $q$. Using brackets $\langle \rangle$ to denote taking the average, let us write $\mathrm{var}\, q_\Sigma = \langle (q^\Sigma - q)^2 \rangle$ and $\mathrm{var}\, q_{\Sigma\chi} = \langle (q^{\Sigma\chi} - q)^2 \rangle$ to denote the variances of firing rate (sigma) and co-firing rate (sigma-chi) decoding errors for $q$, respectively (note that superscripts are used to denote raw estimates for $q$ generated by the decoders, whereas subscripts are introduced to denote the decoding errors of these estimates). We postulate that the trade-off in decoding error between the two channels obeys a relation similar to the uncertainty principle:

$$\mathrm{var}\, q_\Sigma \, \mathrm{var}\, q_{\Sigma\chi} \geq C_q \qquad \text{(Eq. 19)}$$

where $C_q$ denotes a Cramer-Rao bound for estimating $q$ from both decoders together. This equation expresses our postulate that a single stimulus $q$ cannot be more accurately encoded by firing rates and co-firing rates together than by either firing rates or co-firing rates alone. Evidence for this postulate is provided in simulations of Fig. 12, where it was shown that adding position information to the firing rate channel caused a reduction the amount of position information conveyed by the co-firing rate channel, as Eq. 19 would predict.

*Uncertainty trade-off within coding channels*

Now, let us now consider cases where **S** encodes information not only about $q$, but also about its conjugate, $\dot{q} = dq/dt$ (for example, position and velocity). Since $q$ and $\dot{q}$ are conjugate variables, we postulate that the more information either of the two coding channels conveys about $q$, the less information that same channel can convey about $\dot{q}$, and vice versa. For the firing rate channel, this postulated uncertainty trade off may be written as

$$\mathrm{var}\, q_\Sigma \, \mathrm{var}\, \dot{q}_\Sigma \geq C_\Sigma \qquad \text{(Eq. 20)}$$

where $C_\Sigma$ denotes a Cramer-Rao bound for the sigma decoder. Evidence for this postulate is provided by the simulations of Fig. 12, where it was shown that adding position information to the firing rate channel caused a reduction in the amount of velocity information conveyed by the firing rate channel, as predicted by Eq. 20. For the co-firing rate channel, the same within-channel trade-off may be written as

$$\text{var } q_{\Sigma\chi} \text{ var } \dot{q}_{\Sigma\chi} \geq C_{\Sigma\chi} \qquad\qquad (\text{Eq. 21})$$

where $C_{\Sigma\chi}$ denotes a Cramer-Rao bound for the sigma-chi decoder. Evidence for this postulate is again provided by simulations of Fig. 12, where it was shown that a reduction in the amount of position information in the co-firing rate channel was accompanied by an increase in the amount of velocity information conveyed by the co-firing rate channel, consistent with Eq. 21.

*Conjugate coding of conjugate variables*

We have proposed that that if one coding channel conveys information about $q$, then this must come at the expense of conveying information about $q$ in the other channel (Eq. 19), and must also come at the expense of conveying information about $\dot{q}$ withing the same channel (Eqs 20 and 21). But does conveying information about $q$ within one channel come at the expense of conveying information about $\dot{q}$ in the other channel? We postulate that the answer is *no*, because the conjugacy between $R$ and $\dot{R}$ mirrors the conjugacy between $q$ and $\dot{q}$. Consequently, conveying information about $q$ in one channel does not consume the other channel's capacity to convey information about $\dot{q}$. In summary, two conjugate variables only compete with one another for coding capacity within conjugate coding channels, but do not compete with one another for capacity between conjugate coding channels. In the Discussion, we shall explore how this conjugate coding principle might yield new insights into how position and velocity information are represented by neural populations in the hippocampus, how real biological neurons extract information from their spike train inputs, and what role oscillations may play in neural computations that process information about conjugate variables.

**DISCUSSION**

A neural code maps states of the world onto states of the brain, and thus implements a function of the form $W \to B$, where $W$ is a domain of world states, and $B$ is a range of brain states. In simulations above, we showed that a population of spiking neurons can simultaneously convey information via two coding channels: a *firing rate code* ($R$) conveyed by within-cell spike intervals, and a *co-firing rate code* ($\dot{R}$) conveyed by between-cell spike intervals (that is, correlations among spike trains). Together, these two coding channels define a representational space, $R \times \dot{R} \in B$, that contains a larger number of states than $R$ or $\dot{R}$ alone, which in turn suggests that firing rates and spike correlations might together possess a greater capacity to convey information than either channel alone. Consistent with this idea, empirical evidence shows that decoding a rodent's position from spike rates and phases of hippocampal place cells is more accurate than decoding from rate or phase alone (Jensen and Lisman, 2000), and head direction can be more accurately decoded from both the firing rates and spike correlations of HD cells than from either source alone (Peyrache et al., 2015). However, $R$ and $\dot{R}$ are not independent coding channels, since almost any change to a spike train's firing rate will also change its correlation with other spike trains, and thus affect co-firing rates as well. While empirical evidence may show that information can be simultaneously conveyed via firing rates as well as spike correlations, this does not tell us whether the information conveyed via one channel has come *at the expense* of information that might otherwise have been conveyed in the other channel. Firing rates and spike correlations are "entangled" with one another, and therefore, it is not obvious whether more information can be encoded by both channels together than by either channel alone.

## An uncertainty principle for neural coding

Here, we have proposed an uncertainty principle for neural coding which postulates precise conditions under which more information can (or cannot) be encoded by firing rates and co-firing rates together than by either alone. This principle states that $R$ and $\dot{R}$ behave as conjugate coding channels, and therefore, information conveyed via firing rates normally comes at the expense of conveying information via co-firing rates, and vice versa (Eq. 19). If spike trains are used to encode a single variable, $q \in W$, or to encode multiple variables that are not conjugates of one another, $q_1 \times q_2 \times \cdots \times q_N \in W$, then firing rates and co-firing rates together generally cannot convey more information than firing rates or co-firing rates alone. But an exception occurs when $R$ and $\dot{R}$ encode a pair of conjugate variables, $q$ and $\dot{q} = dq/dt$. In such cases, the conjugacy between $R$ and $\dot{R}$ mirrors the conjugacy between $q$ and $\dot{q}$, and the full capacity of $R \times \dot{R}$ becomes "liberated" to encode states within the phase space, $q \times \dot{q} \in W$.

We thus postulated that the full capacity of $R$ and $\dot{R}$ becomes available when spike trains encode information about a pair of conjugate variables. Such cases are of broad interest, because almost any computational task that requires the brain to encode a dynamically changing state variable will also require encoding that same variable's time derivative (i.e., its conjugate). Here, we have focused attention upon dual coding of position and velocity in hippocampal systems to support spatial navigation. But position and velocity are encoded by many other sensory and motor systems as well. Other examples of conjugate variable pairs that are encoded by neural circuits include velocity and acceleration, expected reward and prediction error, phase and frequency, and so on. The conjugate coding principle suggests that whenever and wherever populations of spiking neurons encode pairs of conjugate variables in the brain, a single population of neurons can simultaneously distribute information about both variables across firing rates and co-firing rates. As discussed below, some computational problems may become easier and more efficient to solve under this flexibility to choose between alternative representations for $q$ and $\dot{q}$. Recognizing these advantages may help us to better understand why identified neural populations generate specific patterns of neural spiking (such as peak-shaped tuning functions and neural oscillations), how different populations are interconnected to form functional circuits, and how individual neurons integrate spike train inputs to decode information from both firing rates and spike correlations.

### *The dual coding spectrum*

Schemes for dual coding of conjugate variables reside along a spectrum that spans two extreme endpoints. At one endpoint are cases where $q$ is encoded exclusively by firing rates ($q \rightarrow R$) and $\dot{q}$ is encoded exclusively by co-firing rates ($\dot{q} \rightarrow \dot{R}$). Peak-shaped tuning of firing rates for $q$ is the core mechanism by which this is achieved. For example, in simulations of HD and grid cells with Poisson firing rates, we showed that peak-shaped tuning of firing rates for position ($q$) automatically embeds information about velocity ($\dot{q}$) into co-firing rates (see Figs. 5 & 7). A similar coding scheme could be used to convey information about other pairs of conjugate variables, such as velocity and acceleration, or expected reward and prediction error. For example, encoding acceleration ($\dot{q}$) in co-firing rates could be achieved if firing rates exhibit peak-shaped tuning for velocity ($q$), whereas encoding prediction error ($\dot{q}$) in co-firing rates would require firing rates to exhibit peak-shaped tuning for expected reward ($q$).

At the other extreme endpoint of the dual coding spectrum are cases where $\dot{q}$ is encoded exclusively by firing rates ($\dot{q} \rightarrow R$) and $\dot{q}$ is encoded exclusively by co-firing rates ($q \rightarrow \dot{R}$). The core mechanism for constructing such a code is for firing rates of individual neurons to exhibit linear tuning for $\dot{q}$ in conjunction with peak-shaped tuning for oscillatory phase, which causes information about $q$ and $\dot{q}$ to become packaged into co-firing rates and firing rates, respectively. The role of oscillatory modulation in constructing such codes is not to impose a precise pattern of temporal firing on the spikes (co-firing rate codes are not time codes, as explained in the introduction), but rather to endow co-firing rates with the correct sign (positive or negative) under the coding scheme that they implement (because unlike firing rates, co-firing rates are signed quantities). The period of oscillatory modulation thus sets a time scale for sign reversal of co-firing rates. Coding of position by co-firing rates was demonstrated in simulations of theta-modulated speed cells (Figs. 8-10), but here again, a similar coding scheme could be used to convey information about other pairs of conjugate variables, such as velocity and acceleration, or expected reward and prediction error. For example, encoding velocity ($q$) in co-firing rates would require acceleration ($\dot{q}$) to linearly modulate firing rate differences between oscillatory neurons, and encoding expected reward ($q$) in co-firing rates would require prediction error ($\dot{q}$) to linearly modulate firing rate differences between oscillatory neurons.

In between the two extremes are intermediate cases of dual coding where both $R$ and $\dot{R}$ convey partial information about both variables, $q$ and $\dot{q}$. In such cases, whatever leftover capacity in $R$ not used to encode $q$ can be used to encode $\dot{q}$, and likewise for $\dot{R}$. So spike trains simultaneously implement two mappings, $q \times \dot{q} \rightarrow R$ and $q \times \dot{q} \rightarrow \dot{R}$, as demonstrated by simulations of theta-modulated grid cells (Fig. 12). This kind of intermediate coding for position and velocity requires neurons that simultaneously exhibit peak-shaped tuning for position (to map $q \rightarrow R$ and $\dot{q} \rightarrow \dot{R}$) in conjunction with oscillatory modulation and linear velocity dependence (to map $\dot{q} \rightarrow R$ and $q \rightarrow \dot{R}$). Neural populations in the hippocampal system often exhibit all three of these influences at the same time—position tuning, velocity sensitivity, and oscillatory modulation—suggesting that they do not occupy extreme endpoints of the dual coding spectrum, but rather lie somewhere in between. Some neural populations might lie nearer to one end of the spectrum, such place and grid cells that are strongly modulated by position but weakly modulated by theta oscillations, while other populations might lie near the opposite end of the spectrum, such as theta cells that are weakly modulated by position and more strongly modulated by theta oscillations and running speed. But rather than encoding different variables like position versus speed, as is commonly assumed, these different populations may encode the exactly same pair of conjugate variables (position and velocity) in different ways, by differentially distributing information across firing rates and spike correlations. Why should different populations redundantly encode the same information in different ways? Perhaps because different computations can be performed more efficiently at different points along the dual coding spectrum, so that the flexibility to represent the same information at different points along the spectrum allows a wider range of computations to be optimized (see below).

*Single-neuron computation*

To model how information can be extracted from $R$ and $\dot{R}$, we introduced two biologically inspired decoding methods, *sigma* and *sigma-chi* decoding, which extract information from within- versus between-cell spike intervals, respectively. Sigma decoding is similar to computations performed by model neurons in standard artificial neural networks, which are typically composed of simplified linear units that compute weighted sums of their firing rate inputs, and thus derive their outputs as a function of $R$ (but not of $\dot{R}$). By contrast, sigma-chi decoding requires nonlinear computations to detect correlations among different spike trains, and such computations are not easily reducible to simple weighted linear summation of firing rate inputs (for review, see Mel, 2007). Unlike simple linear neurons in artificial networks, biological neurons have large dendritic trees with detailed branching patterns, endowing them with complex anatomical and electrotonic structure. It has been shown that the dendrites of hippocampal pyramidal cells (Makara & Magee, 2013) and entorhinal stellate cells (Schmidt-Hieber et al., 2017) can perform nonlinear integration of their spike train inputs, which might allow dendrites to perform operations similar to the derivation of chi rates (Eq. 9) that underlie sigma-chi decoding in our simulations. Thus, a single neuron might compute different chi rates in each of its dendrites, and then then pool multiple chi rates together at the soma (Eq. 10) to derive its output as a function of $\dot{R}$ instead of (or in addition to) $R$. A network composed from such sigma-chi neurons would have flexibility to perform computations that operate on representations in $\dot{R}$ as well as $R$. At the circuit level, this flexibility might facilitate novel, efficient solutions to certain computational problems that are unavailable in standard network models composed from linear "sigma" neurons that only operate upon representations in $R$.

*Circuit-level computations*

A wide range of theoretical models have been proposed to explain how neural populations in the hippocampal system—such as place cells, grid cells, border cells HD cells, speed cells, and theta cells—might be functionally interconnected with one another to solve navigational problems such as cognitive mapping, self-localization, and trajectory planning (for review, see Hinman et al, 2018). One type of problem solved such networks is *trajectory planning*, that is, finding the best path through an environment from a start location to a goal. This problem bears resemblance calculating a "path of least action" in classical mechanics, since it involves finding the trajectory that agent should follow as it is influenced by multiple "forces" (goal seeking, obstacle avoidance, effort minimization, etc.). To solve the equations of motion that guide an object along a complex trajectory, physicists rely upon convenient mathematical representations—such as the LaGrangian or Hamiltonian—that cleanly separate the influences of conjugate variables (such as position and momentum) into distinct mathematical terms of a single equation. It is interesting to speculate that neural circuits might employ a similar strategy for simplifying complex trajectory-finding problems, by representing position and velocity in distinct coding channels (firing rates versus co-firing rates) of a single set of neural spike trains. This would be an interesting avenue for future research on dual coding. But for now, we shall focus further discussion upon two other fundamental computations that are commonly performed by neural systems: *differentiation*, which converts position into velocity ($q \rightarrow \dot{q}$), and *integration*, which converts velocity into position ($\dot{q} \rightarrow q$).

## Neural differentiation

We showed that if neurons with peak-shaped positional tuning of their firing rates (such as HD and grid cells) are simulated using Poisson spike trains (Figs. 1-7), then position is encoded exclusively by firing rates ($q \rightarrow R$), and velocity is encoded exclusively by co-firing rates ($\dot{q} \rightarrow \dot{R}$). At this extreme end of the dual coding spectrum, position information is recovered via sigma decoding, whereas velocity information is recovered via sigma-chi decoding. Under these circumstances, sigma-chi decoding can be viewed as tantamount to a process of neural differentiation that recovers $\dot{q}$ from $q$ (e.g., velocity from position). It makes somewhat less sense to conversely view sigma decoding as a process of neural integration that recovers $q$ from $\dot{q}$, because at this end of the dual coding spectrum, the estimate of $\dot{q}$ generated by the sigma-chi decoder is delayed in time with respect to the estimate of $q$ generated by the sigma-chi decoder (see Fig. 2). Therefore, under the reasonable assumption that a computation's input precedes its output in time, it makes more sense to view $q$ as an input and $\dot{q}$ as an output when $q$ precedes $\dot{q}$, and thus to regard sigma-chi decoding as a neural differentiation process. As will be seen below (see "neural integration"), the temporal order of $q$ and $\dot{q}$ reverses at the opposite end of the dual coding spectrum, so that sigma-chi decoding then becomes tantamount to integration rather than differentiation.

### Pre- versus post-positional velocity signals

We use the term "post-positional velocity signals" to describe speed or velocity signals that are derived by differentiating position signals (e.g., via sigma-chi decoding). This is to draw a distinction with "pre-positional" velocity signals that could be derived from the motor or vestibular systems, independently of any positional code. There is evidence that pre- and post-positional velocity signals co-exist in the hippocampal system: some speed cells are prospectively correlated with the animal's future running speed, while others are retrospectively correlated with the past running speed (Kropf et al., 2015). If post-positional speed cells derive their speed tuning via differentiation of position signals, then they should lose their speed modulation following any disruption of the position signal that they differentiate. For example, if retrospective speed cells derive their tuning by differentiating inputs from grid cells (as in simulations of Fig. 7E), then disrupting grid cells should disrupt these retrospective speed cells. Medial septum inactivation has been shown to impair grid cell firing (Brandon et al., 2011; Koenig et al. 2011) while sparing speed cell firing (Hinman et al., 2016) in entorhinal cortex; however, entorhinal speed cells are biased toward prospective coding (Kropf et al., 2015), and may thus not derive their speed tuning by differentiating position signals. By contrast, hippocampal speed cells tend to show a bias for retrospective speed coding (Kropf et al., 2015), so these cells might be more likely to lose their speed tuning after place or grid cells are disrupted. To our knowledge, this has not yet been experimentally tested, but one testable prediction of the conjugate coding hypothesis is that some retrospective speed cells should show impaired speed tuning after disruption of the position signals that they differentiate.

Why should there be separate populations of prospective and retrospective speed cells? Some neural computations in the hippocampal system might depend upon both pre- and post-positional velocity signals. For example, it has recently been shown from place cell recordings that the gain of the hippocampal path integration can be modified by prolonged exposure to a cue conflict between inertial and non-inertial self-motion cues (Jayakumar et al., 2019). Such cue conflicts may generate error signals that recalibrate the gain of the path integrator. One way to compute such error signals would be to calculate the difference between pre- versus post-positional velocity signals, and adjust the gain in proportion to the mismatch. Hence, deriving post-positional velocity signals via sigma-chi decoding might be essential for calibrating the gain of neural integrators in the hippocampal system and other brain systems.

*Differentiation by cells versus circuits*

The fact that velocity information can be extracted from between-cell spike intervals of position-tuned neurons is not a new idea. A similar idea has long been exploited in classical models of visual motion sensitivity, where neurons tuned for the direction of visual motion can be modeled by circuits that differentiate inputs from position-tuned neurons with receptive fields that are arranged sequentially along the preferred path of motion (Hubel & Weisel, 1962, Baccus et al., 2008). Such models bear a strong resemblance to our simulations of angular velocity cells (Fig. 4C) and speed cells (Fig. 7E), in which sigma-chi neurons derive velocity signals from position-tuned neurons. But in classical models of visual motion detection, directional tuning is typically derived from circuit-level mechanisms (such as asymmetric lateral inhibition), whereas in our simulations, velocity tuning was instead achieved by the chi operation (Eq. 9), which could performed at the cellular level (rather than the circuit level) in dendrites of biological neurons (see below). Indeed, calcium imaging experiments have revealed that in visual cortex, most neurons that receive input from orientation-tuned cells pool their inputs across the entire 360° range of edge orientations, and exhibit "hot spots" for specific orientations in different parts of their dendritic tree (Jia et al., 2010). This arrangement is similar to what would be expected if visual cortex neurons were computing co-firing rates by pooling their inputs across multiple orientation phases, in a manner similar to the way that sigma-chi neurons pool inputs across multiple phases of head angle (Fig. 4) or theta phase (Fig. 10) in our simulations.

**Neural integration**

Neural integration plays important roles in spatial coding. Spatially tuned neurons are thought to compute an animal's position in two ways: by measuring the animal's displacement from fixed landmarks in the surrounding environment, and by measuring the animal's current position relative to its prior position (McNaughton et al., 1996). The latter process requires integrating the animal's movement velocity over time to compute its position, a procedure known as *path integration.* Extracting position information via sigma-chi decoding (as in simulations of Figs. 8-10) is tantamount to path integration, since it is a process that derives position ($q$) from velocity ($\dot{q}$).  Three types of models have been proposed to explain how hippocampal networks might perform path integration: attractor networks, reservoir computing models, and oscillatory interference models. It is worthwhile to consider how dual coding of conjugate variables might impact mechanisms of path integration in each of these three classes of models.

*Attractor networks*

Attractor networks have been used to simulate path integration by HD cells (Zhang, 1996; Song & Wang, 2005), place cells (Samsonovich and McNaughton, 1997; Conklin & Eliasmith, 2005; Hedrick & Zhang, 2016), and grid cells (Fuhs & Touretzky, 2006; Guanella and Kiper, 2007; Burak & Fiete, 2009). Although these models differ in their implementation details, they all share two core features of attractor networks in common (see Knierim & Zhang, 2012). First, a population of neurons are reciprocally interconnected with one another via lateral inhibition, causing a localized "activity bump" to form as a stable attractor state of the network that endows individual neurons (such as HD, place, or grid cells) with peak-shaped positional tuning of their firing rates. Second, the symmetry of lateral connections among position-tuned neurons is controlled by neurons with velocity-tuned firing rates, which can push the activity bump through the network along trajectories that mirror the animal's trajectory through space.

Analysis of spike trains from pairs of simultaneously recorded HD cells (Peyrache et al., 2015; Butler & Taube, 2017) and grid cells (Yoon et al., 2013) has revealed evidence for attractor dynamics in these networks, and experimentalists have reported connections in the entorhinal grid cell network that resemble connectivity patterns predicted by standard attractor network models (Couey et al., 2013; Fuchs et al., 2016). However, attractor networks implicitly assume that information about position and velocity is encoded only by firing rates, and not by spike correlations. Consequently, simulated attractor networks are typically composed from linear neurons (either non-spiking units or LIF units) that derive their outputs exclusively from $R$, and not from $\dot{R}$, much like the sigma decoder in our simulations. But what if individual neurons behave more like sigma-chi decoders than sigma decoders? This is a real possibility, as evidenced by the fact that dendrites of hippocampal pyramidal cells and entorhinal stellate cells can perform nonlinear integration of their inputs, and may thus derive their outputs from correlations among their input spike trains rather than linear summation of firing rate inputs (Makara & Magee, 2013; Schmidt-Hieber et al., 2017). The connectivity patterns and neural firing properties predicted by attractor models are founded upon the assumption that individual neurons behave like sigma decoders, but if they instead behave like sigma-chi decoders, then connectivity and firing properties within the hippocampal system could look very different from what standard attractor models predict.

For example, consider a standard attractor network composed from linear units, in which velocity signals that push the activity bump are extracted from firing rates of velocity-tuned neurons. Such a model predicts feedforward connections from velocity-tuned neurons to position-tuned neurons. But under conjugate coding, velocity signals could be extracted from the co-firing rates of position-tuned neurons, so feedforward velocity inputs predicted by standard attractor networks might be minimal or existent. Instead, they could be replaced (or augmented) by functionally equivalent connections from one population of position tuned neurons to another (e.g., from entorhinal grid cells to hippocampal place cells, or vice versa). Since velocity signals encoded by co-firing rates tend to be delayed in time by the integration time constant (see Fig. 2), a time lag would be introduced into velocity signals extracted from co-firing rates of position-tuned neurons. However, position signals encoded by the firing rates

HD cells and grid cells are often seen to prospectively lead the animal's true position (Muller & Kubie, 1989; Blair & Sharp, 1995; Mehta et al., 1997; Almeida et al., 2012), which could help to compensate for time lag if velocity signals were derived from the co-firing rates of these neurons for the purpose of path integration. Standard attractor networks also predict lateral inputs to position tuned neurons from neighboring position tuned neurons, but under conjugate coding, these connections too could be replaced or augmented by functionally equivalent alternatives, such as inputs to position tuned neurons (e.g., place cells or grid cells) from velocity-tuned neurons (e.g., theta cells) that encode position in their co-firing rates rather than their firing rates. In summary, the conjugate coding principle suggests that biological and artificial attractor networks may be subject to radically different constraints upon their connectivity, and therefore, attractor models composed purely from linear neurons may be quite limited in their ability generate accurate predictions about connectivity in biological path integration networks.

Dual coding of position and velocity might also help to explain why some populations of neurons that are predicted to exist by standard attractor models have never been observed. In two-dimensional open-field environments, place cells and grid cells exhibit robust positional tuning of their firing rates in two dimensions. Path integration in two dimensions requires a two-dimensional representation of movement velocity, but to our knowledge, no experiment has ever observed neurons with firing rates that are tuned for velocity in two dimensions. Speed cells may contribute to encoding the non-directional component of velocity in two dimensions, but an additional component for movement direction would also be required. The entorhinal cortex does contain neurons that are tuned for head direction (Giocomo et al., 2014), and many grid cells are selective not only for the animal's position but also for the direction in which its head is facing (Sargolini et al., 2006; Wills et al., 2012). However, these entorhinal populations appear to be tuned for the head's azimuthul position, rather than for azimuthal movement direction. Accurate two dimensional path integration requires an azimuthal velocity signal, rather than an azimuthal position signal (Roudies et al., 2015). If velocity is encoded in spike correlations as well as (or instead of) firing rates, then a firing rate code for azimuthal movement velocity would not need to exist, because neurons that encode a two-dimensional position signal in their firing rates (such as place or grid cells) would automatically also encode a two-dimensional velocity in their co-firing rates. Hence, the co-firing rates of grid cells could serve as the substrate for encoding velocity signals that support two-dimensional path integration, even in the absence of any neural population that encodes two-dimensional velocity signals in their firing rates. A similar idea has been proposed by Zutshi et al. (2017), who argued that theta sequences generated by grid cells may provide a substrate for two-dimensional velocity coding in entorhinal cortex. However, our simulations suggest that while oscillatory modulation (which gives rise to theta sequences) is essential for encoding position signals in spike correlations, it is not essential for encoding velocity signals in spike correlations, as demonstrated by the fact that velocity can be recovered via sigma-chi decoding from simulated HD cells (Fig. 5) and grid cells (Fig. 7) even when they generate Poisson spike trains with no oscillatory modulation at all.

*Reservoir computing networks*

Reservoir computing models of path integration are recurrent neural networks trained from example data (via gradient descent methods) to convert time-varying velocity signals into time-varying position signals (Abbott et al., 2016, DeNeve & Machens, 2016). It has recently been shown that neurons with periodic spatial tuning—similar to entorhinal grid cells—can emerge spontaneously in a non-spiking recurrent network trained to perform spatial path integration (Banino et al., 2018). Like attractor models, most reservoir computing models of path integration use recurrent networks composed from linear neurons, so the inputs and outputs to these networks encode velocity and position signals solely as vectors of neural firing rates (not spike train correlations). However, unlike standard attractor models, recurrent connections within a reservoir computing network are sculpted by learning. It is thus possible that when a recurrent network of spiking neurons is trained to perform path integration, error-driven learning could cause the formation of neural microcircuits that extract information about position or velocity from correlations among spike trains, and not just from firing rates. This would be more likely to occur if the network were constructed not from  linear neurons, but from spiking neurons that perform nonlinear integration of their inputs, similar to the sigma-chi neurons in our simulations. Training a recurrent network of sigma-chi neurons to perform path integration would require differentiating Eq. 9 to derive a cost function that supports convergent learning. If this could be achieved, then it would be intriguing to investigate what kinds of coding mechanisms naturally emerge when a recurrent network of sigma-chi neurons is trained to perform path integration. Phenomena such as oscillatory rhythms that support dial coding might emerge spontaneously from the training of such a network, in a manner similar to the way that grid cells emerge spontaneously when a network of linear neurons is trained to perform path integration (Banino et al., 2018). The emergent firing properties and connectivity patterns in a trained network of recurrently connected sigma-chi neurons might bear closer resemblance to real hippocampal and entorhinal networks than those that emerge from networks of linear neurons.

*Oscillatory interference models*

Oscillatory interference models are explicitly designed to encode position information using correlated neural activity, rather than firing rates. These models propose that the brain contains *velocity-controlled oscillators* (VCOs) that shift phase against one another at a rate that depends upon the animal's movement velocity (Burgess et al., 2007; Geisler et al., 2007). Consequently, phase offsets between oscillators depend upon the animal's positon in space. If VCOs are implemented by spike trains of neurons that burst rhythmically at the theta frequency, then the VCOs map the animal's position into a representational space where each dimension measures time intervals (normalized by the theta cycle period) between pairs of spikes that are fired by *different* neurons (that is, cells synchronized to different VCOs). This is in marked contrast with attractor and reservoir computing models, which map the animal's position into a representational space where each dimension measures time intervals between pairs of spikes that are fired by the *same* neuron (that is, firing rates).

Early oscillatory interference models demonstrated how individual grid cells (Burgess et al., 2007; Giocomo et al., 2007a,b) or place cells (Blair et al., 2008) could derive their position-tuned firing rates by detecting of location-specific synchrony among inputs from theta VCOs. This mechanism can account not only for the spatial firing properties of place and grid cells, but also for temporal firing properties such as theta rhythmicity, phase precession (O'Keefe & Recce, 1993; Hafting et al., 2008), and modulation of theta oscillations by running speed (Geisler et al., 2007; Welday et al., 2011; Jeewajee et al., 2014). Some of these early oscillatory interference models (Burgess et al., 2007; Giocomo et al., 2007a,b) simulated nonlinear multiplicative interactions among VCO inputs to grid cells, while others were based upon linear summation of VCO inputs (Welday et al., 2011). However, these models were focused upon mimicking the firing properties of individual grid or place cells, rather than modeling path integration or other network-level computations. Later models proposed novel "hybrid" architectures for performing path integration through a combination of both attractor dynamics and oscillatory interference (Bush & Burgess, 2014; Hasselmo & Shay, 2014). Despite incorporating oscillatory interference mechanisms rooted in spike correlations, these models were simulated by networks of linear neurons that derived their outputs by computing weighted sums of their inputs, and then converted these weighted sums into spike trains using LIF or Izhikevic spike dynamics. The connectivity patterns predicted by these models might change— and perhaps become more biologically accurate—if they were reformulated using model neurons that perform nonlinear integration of their inputs to take full advantage of dual coding by firing rates as well as spike correlations.

**Summary and Conclusions**

Neuroscience researchers are sometimes prone to "ratism"—a bias to regard the firing rates of spiking neurons as the primary coordinate basis of the neural code. Under the influence of this bias, it is reflexively assumed that if individual neurons are tuned for some particular variable (such as head angle, spatial position, or running speed), then a population of such neurons exists mainly to encode distributed representations of that same variable using vectors of neural firing rates. While it is certainly true that information can be encoded by (and decoded from) population vectors of neural firing rates, it is does not follow from this that the space of firing rate vectors is the only representational space into which information is mapped by neural spike trains. It is necessary to recognize and appreciate that the same population of neurons can encode different information when viewed through the "lenses" of different decoders.

Here, we have shown that there is a conjugate relationship between firing rate codes and spike correlation codes, which mirrors the uncertainty principle from physics. A firing rate code for position ($q \rightarrow R$) can co-exist with a co-firing rate code for velocity ($\dot{q} \rightarrow \dot{R}$), or a co-firing rate code for position ($q \rightarrow \dot{R}$) can co-exist with a firing rate code for velocity ($\dot{q} \rightarrow R$). More generally, a firing rate code for any time-varying stimulus can co-exist with a co-firing rate code for the time derivative of that stimulus, and vice versa. Hence, the conjugate coding principle described here may be useful for understanding neural coding not only in the hippocampal system, but in other brain systems as well. For example, the visual system encodes information about the position and as well as the velocity of objects in the visual field. Motor and proprioceptive systems processes information about the position and velocity of moving limbs. And neural circuits for reinforcement learning encode information about expected value and prediction error, which bear a relationship to one another that is similar to the relationship between position and velocity (prediction error is the time derivative of expected value, and expected value is the time integral prediction error). In all of these systems, the conjugate coding principle may be at work to embed orthogonal representations of stimuli and their time derivatives within the firing rates and co-firing rates of neural spike trains. Further theoretical and empirical analysis may help to elucidate how biological neurons extract information (at the level of single cells) from both firing rates and co-firing rates to efficiently perform useful computations.

**Acknowledgements**

We thank Kechen Zhang for valuable discussions about how to mathematically formulate our ideas about the uncertainty principle and dual coding of conjugate variables. Manu Madhav and Ravi Jayakumar provided position tracking data from the circular track, and Rose DeGuzman assisted in collecting the head direction data from the open field. We also thank Joseph Monaco for helpful feedback and discussion. This work was supported by NIH Initiative for Maximizing Student Development (IMSD) grant #GM055052 to Ryan Grgrich.

**METHODS**

All simulations were carried out using MATLAB (source code available upon request).

*Position Tracking*

Simulated HD, grid, and theta cells were generated from position tracking data. Head direction data from the open field was obtained while a rat foraged freely for food pellets in a circular arena (diameter = 80 cm).  Rats wore a pair of red and green light-emitting diodes (LEDs) spaced 11.25 cm apart from one another, and an overhead video camera sampled LED positions at $R$=30 Hz with a resolution of $P$=4.7 pixels/cm. Each LED's position was smoothed using a boxcar window 15 samples (0.5 s) wide, and the rat's head direction at each time step was estimated as $\arctan(\Delta y_i/\Delta x_i)$, where $\Delta x_i$ and $\Delta y_i$ denote the difference between the $x$ and $y$ coordinates of the red and green LEDs, respectively a the $i^{th}$ position sample. Position data from the circular track was obtained while rats ran laps on a 1.5 m diameter circular track enclosed within a planetarium-style dome, where an array of three visual landmarks was projected onto the interior surface to create an augmented reality environment (Jayakumar et al., 2019). The rat was attached to a boom arm that rotated around a joint in the center of the track, and the rat's position on the track was sampled at 100 Hz from an optical encoder of the boom angle in the center joint.

*Generation of simulated spike trains*

To simulate spike trains of HD cells or grid cells, we computed the probability of spiking at each time step using Eq. 3. The computed probability was then used as a threshold for a random number generator that output a real value between 0 and 1; a spike was placed in a given time step if the random value fell below the calculated probability threshold, yielding a binary spike response function (Eq. 1). For simulations of theta cell spike trains, the spike response function was generated in two steps. First we first created a "seed spike train" for each theta cell, containing a single spike at every time step where the theta cell's phase (computed using Eq. 16) passed through perfect phase synchrony with the reference oscillator. The probability of spiking for theta cell $n$ was then computed by convolving the seed spike train with a Gaussian kernel (Eq. 18). A problem with this methods was that when seed spikes were spaced closer together than the width of the Gaussian kernel, it was possible for the convolution procedure to yield inappropriately large spike probabilities. To prevent this, the convolution was performed separately on even and odd numbered spikes within the seed spike train, so that no two spikes would be closer together than the width of the Gaussian kernel. The two convolution results were then merged by taking the maximum probability from either of the two results at each time step. The probabilities in the merged time series were then used as the probability threshold on the random number generator for outputting the theta cell's spike response function.

*Conversion of speed cells to grid cells*

Simulated speed cell spike trains were converted to simulated grid cell spike trains (Fig. 12) in the following way. Each speed cell resided at one of twelve positions in a ring oscillator, so we may index speed cells by the variable $i = 1, 2, \cdots 12$. The conversion process introduced a spatially periodic firing rate modulation into the speed cell's spike train, converting it into a grid cell. The probability of position-driven spiking under this modulatory influence was given by $\rho_i(q) = \cos(3q + 2\pi i/12)$, where $q$ is the animal's angular position on the track, 3 cycles per lap is the spatial frequency of the periodic modulation, and $i$ shifts the spatial phase of the grid field. At all time steps where speed cell $i$ fired a spike and $\rho_i$<0, a random number $r_i(t) \in \{0,1\}$ was generated and the spike was deleted if $r_i(t) < |\rho_i|$. At all time steps where speed cell $i$ did not fire a spike and $\rho_i$>0, a random number $r_i(t) \in \{0,1\}$ was generated and a spike was added if $r_i(t) < \vartheta|\rho_i|$, where $\vartheta = .0475$ was a coefficient that was empirically determined to leave the mean number of spikes fired by the speed cell unchanged after it had been converted to a grid cell, using the speed cell simulation parameters from the simulations in Fig. 11.